\documentclass{article}

\newif\ifarxiv
\arxivtrue
\newif\ifcomments
\commentsfalse
\iftrue
\ifarxiv
\usepackage[accepted]{icml2022}
\else
\usepackage{icml2022_style/icml2022}
\fi
\fi
\usepackage{hyperref}
\usepackage[utf8]{inputenc}
\usepackage{amsmath}
\usepackage{amstext}
\usepackage{amsthm}
\usepackage{amssymb}
\usepackage{mathtools} 
\usepackage{algorithmic}
\usepackage{algorithm}
\usepackage{xcolor}

\usepackage[noabbrev,capitalize]{cleveref}

\hypersetup{
  colorlinks = true,
  citecolor = blue,
}

\DeclarePairedDelimiter{\parens}{(}{)}
\usepackage[normalem]{ulem}
\newcommand{\realLine}{\mathbb{R}}
\newcommand{\targetDimensionality}{M}
\newcommand{\genericIntegrand}{f}
\newcommand{\KL}{\textup{KL}}
\newcommand{\G}{G}
\newcommand{\overallObjective}{H}
\newcommand{\targetDist}{\pi}

\newcommand{\forwardGenerating}{\bar{\eta}}
\newcommand{\augmentedTarget}{\bar{\targetDist}}
\newcommand{\unnormDensity}{\gamma}
\newcommand{\baseFlow}{T}
\newcommand{\genericFlowParams}{\theta}
\newcommand{\klTargetDist}{\mu_{\timeIndex}}

\newcommand{\markovTransitionKernel}{\mathcal{K}}
\newcommand{\markovTransitionKernelReverse}{\tilde{\markovTransitionKernel}}

\newcommand{\expect}{\mathbb{E}}
\newcommand{\x}{x}
\newcommand{\rvX}{X}
\newcommand{\rvY}{Y}
\newcommand{\particleX}{\rvX}
\newcommand{\particleY}{\rvY}

\newcommand{\normConst}{Z}
\newcommand{\numTransitions}{K}
\newcommand{\timeIndex}{k}
\newcommand{\genericGroup}{\mathcal{G}}
\newcommand{\genericGroupElement}{g}
\newcommand{\flowTargetDensity}{\pi_{T}}
\newcommand{\flowBaseDensity}{\pi_{B}}
\newcommand{\genericGroupRepresentationElement}{D}
\newcommand{\particleIndex}{i}
\newcommand{\particleIndexB}{j}

\newcommand{\particleWeightUnnorm}{w}
\newcommand{\particleWeightNorm}{W}
\newcommand{\numParticles}{N}
\newcommand{\pushForward}{\#}
\newcommand{\resamplingThreshold}{A}
\newcommand{\initialDensity}{\pi_0}
\newcommand{\finalUDensity}{\gamma_1}
\newcommand{\annealedUDensity}{\gamma_{\beta}}
\newcommand{\annealedDensity}{\pi_\beta}

\newcommand{\numCraftIters}{J}
\newcommand{\craftIterIndex}{j}
\newcommand{\discreteNeighbour}{\zeta}

\newcommand{\mass}{m}
\newcommand{\coupling}{\lambda}

\newcommand{\latticeSizeX}{x}
\newcommand{\latticeSizeY}{y}
\newcommand{\action}{S}
\newcommand{\latticeX}{\hat{x}}
\newcommand{\latticeUnitVector}{\hat{e}}
\newcommand{\latticeDirection}{\mu}
\newcommand{\deltaDensity}{D}
\newcommand{\braces}[1]{\left\lbrace #1 \right\rbrace}

\newtheorem{prop}{Proposition}

\usepackage{tikz}
\usetikzlibrary{decorations.pathmorphing} %
\usetikzlibrary{matrix} %
\usetikzlibrary{arrows} %
\usetikzlibrary{calc} %

\tikzstyle{block} = [draw,rectangle,thick,minimum height=2em,minimum width=2em]
\tikzstyle{sum} = [draw,circle,inner sep=0mm,minimum size=2mm]
\tikzstyle{connector} = [->,thick]
\tikzstyle{line} = [thick]
\tikzstyle{branch} = [circle,inner sep=0pt,minimum size=1mm,fill=black,draw=black]
\tikzstyle{guide} = []
\tikzstyle{snakeline} = [connector, decorate, decoration={pre length=0.2cm,
                         post length=0.2cm, snake, amplitude=.4mm,
                         segment length=2mm},thick, magenta, ->]

\icmltitlerunning{Continual Repeated Annealed Flow Transport Monte Carlo}

\begin{document}

\twocolumn[ 
\icmltitle{Continual Repeated Annealed Flow Transport Monte Carlo}

\begin{icmlauthorlist}
\icmlauthor{Alexander G. D. G. Matthews}{DeepMind}
\icmlauthor{Michael Arbel}{INRIA}
\icmlauthor{Danilo J. Rezende}{DeepMind}
\icmlauthor{Arnaud Doucet}{DeepMind}
\end{icmlauthorlist}

\icmlaffiliation{INRIA}{Universit\'e Grenoble Alpes, Inria, CNRS}
\icmlaffiliation{DeepMind}{DeepMind}
\icmlcorrespondingauthor{Alexander G. D. G. Matthews}{alexmatthews@google.com}
\icmlcorrespondingauthor{Arnaud Doucet}{arnauddoucet@google.com}

\icmlkeywords{Machine Learning, ICML}

\vskip 0.3in
]
\printAffiliationsAndNotice{}

\begin{abstract}
We propose Continual Repeated Annealed Flow Transport Monte Carlo (CRAFT), a method that combines a sequential Monte Carlo (SMC) sampler (itself a generalization of Annealed Importance Sampling) with variational inference using normalizing flows. The normalizing flows are directly trained to transport between annealing temperatures using a KL divergence for each transition. This optimization objective is itself estimated using the normalizing flow/SMC approximation. We show conceptually and using multiple empirical examples that CRAFT improves on Annealed Flow Transport Monte Carlo \citep{Arbel2021}, on which it builds and also on Markov chain Monte Carlo (MCMC) based Stochastic Normalizing Flows \citep{wunoe2020stochastic}. By incorporating CRAFT within particle MCMC, we show that such learnt samplers can achieve impressively accurate results on a challenging lattice field theory example.
\end{abstract}

\section{Introduction}

There are few algorithmic problems richer or more fundamental than the task of drawing samples from an unnormalized distribution and approximating its normalizing constant. In this paper we combine two important methods for this task specifically Sequential Monte Carlo (SMC) \citep{Del-Moral:2006} and variational inference with Normalizing flows (NFs) \citep{Rezende:2015}.

The first of these components, SMC samplers, are an importance sampling based algorithm.
They are a principled way to combine annealing, resampling and Markov chain Monte Carlo (MCMC) and as such enjoy continuing popularity \citep{dai2020invitation}. One potential disadvantage of SMC samplers is that at each successive temperature step in a SMC sampler there is importance sampling- albeit between adjacent distributions. For high dimensional target distributions this can lead either to estimators with high error or require many temperatures. Another potential disadvantage is that for finite particle numbers, SMC estimates of expectation w.r.t. the target distribution are biased. This can be mitigated by using the SMC sampler inside a Particle MCMC outer loop \citep{Andrieu2011} at the cost of adding repeated SMC sampler calls. It is particularly desirable in this Particle MCMC context for the base SMC sampler to be fast and output low variance estimates of the normalizing constant.

The other component of our method is variational inference with NFs. Flow methods learn a differentiable invertible transformation (diffeomorphism) typically with a tractable Jacobian \citep{papamakarios2019normalizing}. The learnt mappings are both flexible and fast to evaluate but there are still some challenges around using flows. The tractability of training with reverse Kullback--Leibler divergence comes at the cost of its well known mode seeking behaviour. Further placing the full burden of modelling on a parameterized flow can lead to high memory usage in contrast to non-parametric sampling algorithms. Finally, as diffeomorphisms, flows preserve topological properties of their input. This property can represent a challenge whenever there is topological mismatch between the, typically, simple base distribution and the complex target \citep{Cornish2020}.

Methods combining annealed samplers with normalizing flows have the potential to fix the limitations of each component part. The use of flows can reduce the variance from importance sampling. The MCMC steps can reduce the topological and representational burden on the flows. Used correctly, annealing can reduce the mode seeking behaviour of the variational objective. Resampling can be used to focus computation on promising samples. The idea of combining \emph{fixed} transformations with annealing goes back as far as \citep{vaikuntanathan2011escorted}. It is the \emph{training} of the flows that still presents a challenge and the reason that such combined approaches have not met their full potential. Two recent papers target this problem \citep{wunoe2020stochastic, Arbel2021}, but as we shall relate, neither is entirely satisfactory or as well suited to the task as the algorithm we propose. 

CRAFT is a method for estimating an SMC sampler augmented with interleaved normalizing flows. Unlike many other approaches in this area that incorporate sampling ideas, it is able to cope with full MCMC steps with Metropolis accept/reject corrections and SMC resampling steps. The CRAFT objective is not a standard variational bound. Rather it uses a KL divergence for each transition between temperatures. We find that as well as outperforming other methods for estimating the flows in practice (Section  \ref{section:CRAFTvsAFT} and Section \ref{section:CRAFTvsSNF}), our learnt sampler is useful as a fast inner loop for a Particle MCMC sampler (Section \ref{section:phiFour}).

\section{Method}
In this section we describe and motivate both the sampling method and how to learn the required flows. Section \ref{section:fixedFlows} describes the method for fixed flows, which is equivalent to an SMC sampler with added fixed normalizing flows. In Section \ref{section:conventional_elbo}, we discuss conventional evidence lower bound (ELBO) methods for learning the flows and their limitations. Finally, we present the CRAFT training method in Section \ref{section:craft_training}, and also explain how we improve over the AFT method on which CRAFT builds, by solving what we call the \emph{sample replenishment problem}.

\subsection{Sampler for fixed flows}\label{section:fixedFlows}

We start by describing the sampler for fixed normalizing flows.
In this case, it corresponds to a generalization of the standard SMC sampler which adds deterministic normalizing flows to the usual steps in SMC. This generalization reduces to a standard SMC sampler when the normalizing flows are the identity. \citet{Arbel2021} give a detailed history. The algorithm is described in Algorithm \ref{algo:smcFlow} which sequentially calls Algorithm \ref{algo:SMC_NF}. 

We now describe the steps and derivation in more detail. We consider a sequence of distributions $\left( \targetDist_{\timeIndex}(\x) \right)_{\timeIndex=0}^{\numTransitions}$ on $\mathbb{R}^M$ where $\targetDist_{\timeIndex}(\x) {=}  \frac{\unnormDensity_\timeIndex(\x)}{\normConst_{\timeIndex}}
$ and $\unnormDensity_{\timeIndex}(x)$ can be evaluated pointwise. The final distribution $\targetDist_{\numTransitions}(\x)$ is the distribution with unknown normalizing constant $\normConst_{\numTransitions}$ that we wish to approximate. For the initial distribution $\targetDist_{0}(\x)$ we assume that we can draw exact samples and that the normalizing constant $\normConst_{0}=1$ without loss of generality. 
The intermediate distributions enable us to transition smoothly (or in physics parlance `anneal') from the tractable $\targetDist_{0}(\x)$ to our goal $\targetDist_{\numTransitions}(\x)$. While they will sometimes be of interest themselves, they will often only be used  to help us construct approximations.
The method returns $\numParticles$ weighted particles $\left(\particleX^\particleIndex_\numTransitions, \particleWeightNorm^\particleIndex_\numTransitions \right)_{\particleIndex=1}^{\numParticles}$ which can be used to provide an unbiased estimate of the normalizing constant $\normConst_\numTransitions$ and consistent estimates of expectations under the target $\pi_\numTransitions$.

To build up intuition, consider a forward sampling process $\forwardGenerating$  producing a sequence $(X_\timeIndex)_{\timeIndex=0}^K$ defined on $\realLine^{\targetDimensionality \times (\numTransitions+1)}$ such that the final sample $X_K$ approximates $\targetDist_\numTransitions$. The process starts with an initial sample $\rvX_0$ drawn from some distribution $\targetDist_0$ that is successively transformed using an interleaved sequence of $\numTransitions$ NFs $(\baseFlow_\timeIndex)_{\timeIndex=1}^\numTransitions$ and Markov transition kernels $(\markovTransitionKernel_{\timeIndex})_{\timeIndex=1}^\numTransitions$ with invariant distributions $(\targetDist_\timeIndex)_{\timeIndex=1}^\numTransitions$:

 \begin{tikzpicture}[scale=1., auto, >=stealth']
    \hspace{-0.3cm}
    \small
    \matrix[ampersand replacement=\&, row sep=0.2cm, column sep=0.2cm] {
    
            \node (e0) {};
            \&\&\&\&
      \node[block] (F1) {{ $\begin{aligned}
           Y_{\scalebox{0.85}{1}} &{=} T_{\scalebox{0.85}{1}}(X_{\scalebox{0.85}{0}})\\
           X_{\scalebox{0.85}{1}} &{\sim} \mathcal{K}_{\scalebox{0.85}{1}}(Y_{\scalebox{0.85}{1}},\cdot)
          \end{aligned}$}};
      \&\&\&
            \node (e2) {{$\cdots$}};
       \&\&\&
           \node[block] (f1) {{$\begin{aligned}
           Y_{\scalebox{0.85}{K}} &{=} T_{\scalebox{0.85}{K}}(X_{\scalebox{0.85}{K-1}})\\
           X_{\scalebox{0.85}{K}} &{\sim}  \mathcal{K}_{\scalebox{0.85}{K}}(Y_{\scalebox{0.85}{K}},\cdot)
          \end{aligned}$}}; 
      \&\&\&
      \node (e1) {};

      \&
      \&
      \&
      \\
    };

    \draw [connector] (e0) -- node {{\tiny $X_{\scalebox{0.85}{0}}{\sim}\pi_{\scalebox{0.85}{0}}$}} (F1);
    \draw [connector] (F1) -- node {{\tiny $X_{\scalebox{0.85}{1}}$}} (e2);
    \draw [connector] (e2) -- node {{\tiny $X_{\scalebox{0.85}{K{-}1}}$}} (f1);
    \draw [connector] (f1) -- node {{\tiny $X_{\scalebox{0.85}{K}}$}} (e1);
  \end{tikzpicture}
Intuitively, the purpose of each flow $\baseFlow_\timeIndex$ is to `transport' samples from a density $\targetDist_{\timeIndex-1}$ corresponding to an annealing temperature $\timeIndex{-}1$ towards the next density $\targetDist_\timeIndex$ at temperature $\timeIndex$. 
Successive densities are usually similar, making the flow easier to learn compared to a flow directly transporting $\targetDist_0$ towards the target $\targetDist_\numTransitions$. Since it is generally hard to find a flow $\baseFlow_\timeIndex$ that \emph{perfectly} transports between successive densities, the sampler employs two correction mechanisms:  (1) the Markov transition kernel $\markovTransitionKernel_{\timeIndex}$ with invariant density $\targetDist_\timeIndex$ further diffuses the samples towards $\targetDist_\timeIndex$ and (2) Importance Sampling re-weights the samples to correct for any mismatch between $\targetDist_\timeIndex$ and the density $\baseFlow_\timeIndex^{\pushForward}\targetDist_{\timeIndex-1}$ obtained by transporting $\targetDist_{\timeIndex-1}$ using the flow $\baseFlow_\timeIndex$.

{ \bf  Importance Sampling.} Ultimately, we are interested in re-weighting the final sample $X_K$ to approximate the target $\targetDist_\numTransitions$. 
This could be theoretically achieved by evaluating the target to marginal ratio $\targetDist_\numTransitions/\eta_\numTransitions$, where $\eta_K$ is obtained by integrating the proposal distribution $\forwardGenerating$ over all previous variables $(\rvX_\timeIndex)_{\timeIndex=0}^{\numTransitions-1}$. 
To avoid computing such intractable integrals, we follow the standard approach in SMC and Annealed Importance Sampling (AIS) \citep{neal2001annealed} of computing importance weights between the whole forward process $\forwardGenerating$ and an \emph{augmented target} $\augmentedTarget$ admitting $\targetDist_\numTransitions$ as marginal at time $\numTransitions$. While there are multiple choices for the \emph{augmented target} $\augmentedTarget$, as discussed in \citep{Del-Moral:2006}, 
our choice of $\augmentedTarget$ is made for tractability of the importance weights and corresponds to a formal \emph{backward process} starting by sampling exactly from $\targetDist_\numTransitions$ and then sequentially generating samples backwards in time using the \emph{reversal} transformations of each flow $\baseFlow_\timeIndex$ and MCMC kernel $\markovTransitionKernel_\timeIndex$. 
The reversal of each flow transformation is its inverse while the reversal of each forward Markov kernel is the Markov kernel $\markovTransitionKernelReverse_{\timeIndex}$ satisfying $\targetDist_{\timeIndex}(\x)\markovTransitionKernel_{\timeIndex}(\x,\x') {=}\targetDist_{\timeIndex}(\x') \markovTransitionKernelReverse_{\timeIndex}(\x', \x) $. 
When choosing a reversible Markov kernel for the forward kernel $\markovTransitionKernel_\timeIndex$, %
the reversal kernel is equal to the forward kernel, i.e. $\markovTransitionKernelReverse_{\timeIndex} {=} \markovTransitionKernel_{\timeIndex}$. 
For our choice of augmented target, the \emph{unnormalized} importance weights take the form: 
\begin{subequations}
	\begin{align}
		\particleWeightUnnorm_{\numTransitions}(x_{0:K-1}) &= \prod_{\timeIndex=1}^{\numTransitions} \G_{\timeIndex}(\x_{\timeIndex-1}), \label{eq:aisWeight}\\
		\G_{\timeIndex}(\x_{\timeIndex-1}) &:= \frac{\unnormDensity_{\timeIndex}(\baseFlow_{\timeIndex}(\x_{\timeIndex-1}))}{\unnormDensity_{\timeIndex-1}(\x_{\timeIndex-1})}|\nabla \baseFlow_{\timeIndex}(\x_{\timeIndex-1}) |. \label{eq:GDef}
	\end{align}
\end{subequations}
When the flow is the identity, equation \eqref{eq:GDef} reduces to the ratio of successive densities, which is the standard annealed importance sampling expression. For non-identity flows the $\baseFlow_\timeIndex$ dependent terms correct for the use of the flow. See Appendix \ref{sec:appextendedproposaltargetSNF} for a precise mathematical description of $\forwardGenerating$, $\augmentedTarget$ and the derivation of the weights \eqref{eq:aisWeight}-\eqref{eq:GDef}.

The algorithm is then implemented sequentially to maintain a set of $\numParticles$ particles $(\particleX_{\timeIndex}^{(\particleIndex)}, \particleWeightNorm_{\timeIndex}^{\particleIndex})_{\particleIndex=1}^\numParticles$ consisting in pairs of samples and corresponding normalized IS weights at each time $\timeIndex$ where the samples are obtained using the forward generating process $\forwardGenerating$. The unnormalized IS weights $\particleWeightUnnorm_\timeIndex^\particleIndex$ are computed recursively using $\particleWeightUnnorm_\timeIndex^{\particleIndex} {=} \particleWeightNorm_{\timeIndex-1}^{\particleIndex}\G_\timeIndex\parens{\rvX_{\timeIndex-1}^\particleIndex}$, thus allowing to compute the normalized IS weights $\particleWeightNorm_\timeIndex^{\particleIndex}$ by normalizing over the set of particles $\particleWeightNorm_\timeIndex^{\particleIndex} {=} \particleWeightUnnorm_\timeIndex^{\particleIndex}/\sum_{\particleIndexB=1}^{\numParticles} \particleWeightUnnorm_\timeIndex^\particleIndexB$. 
A consistent approximate expectation of a function $\genericIntegrand$ under the target $\targetDist_\timeIndex$ is then given by $\sum_{\particleIndex=1}^\numParticles \particleWeightNorm_\timeIndex^{\particleIndex} \genericIntegrand(\rvX_\timeIndex^{(\particleIndex)}).$ Moreover, an unbiased estimate $\normConst_\timeIndex^{\numParticles}$ of the normalizing constant $\normConst_\timeIndex$ is obtained sequentially from the previous one $\normConst_{\timeIndex-1}^{\numParticles}$ using the update $ \normConst_\timeIndex^{\numParticles}{=} \normConst_{\timeIndex-1}^{\numParticles}\parens{\sum_{\particleIndex=1}^{\numParticles} \particleWeightUnnorm_\timeIndex^{\particleIndex} }$. 
If the NFs transport perfectly between the temperatures, this algorithm enjoys the property that the normalizing constant estimate $\normConst_\numTransitions^\numParticles$ has a zero variance and the corresponding approximate expectations are unbiased with a variance corresponding to the one of the true target.

{\bf Resampling.} 
Up to this point, we have described AIS with additional fixed normalizing flows. We now go a step further to incorporate resampling, a key ingredient of SMC methods that has proved to be very beneficial 
\citep{Arbel2021,chopin2002sequential,Del-Moral:2006,hukushima2003population}. %
Resampling consists in randomly selecting a particle $\rvX_\timeIndex^{\particleIndex}$ with probability $\particleWeightNorm_\timeIndex^{\particleIndex}$ and repeating the operation $\numParticles$ times to construct a set of $\numParticles$ particles approximately distributed according to $\targetDist_\timeIndex$. 
This operation is equivalent to assigning a number of offspring $N_k^i$ to each particle $X_k^{i}$ and associating a uniform weight of $1/N$ to each offspring. The vector of all offspring $(N_k^i)_{i=1}^N$ is then drawn from a multinomial distribution with weights $\parens{W_k^{i}}_{i=1}^N$ under the constraint that $\sum_{i=1}^N N_k^i {=} N$. The expectation under $\pi_k$ of a function $f$ is then approximated using $\sum_{i=1}^N \frac{N_k^i}{N}f(X_k^{i})$ 
which is also an unbiased estimate of the weighted sum $\sum_{i=1}^N W_k^{i}f(X_k^{i})$. Hence, resampling refocuses computational effort on promising particles (the ones falling in high density regions of $\pi_k$) whilst preserving the key analytic properties of the algorithm.
However, to avoid the additional variance coming from the multinomial distribution, we only use it when the \emph{effective sample size}  $\textup{ESS}_k^N{:=}\parens{\sum_{i=1}^N (W_k^i)^2 }^{-1}$ of the particles  falls beyond some predefined proportion $A{\in}[1/N,1)$ (we use $A=0.3$ in all our experiments) of the total number of samples $N$ \citep{liuchen1995}. Since resampling is followed by the Markov kernel step in each iteration of Algorithm \ref{algo:SMC_NF} any degeneracy introduced during resampling can be reduced.
\begin{algorithm}
\caption{SMC-NF-step}\label{algo:SMC_NF}
	\begin{algorithmic}[1]
		\STATE \textbf{Input:} Approximations $(\targetDist^\numParticles_{\timeIndex-1},\normConst_{\timeIndex-1}^\numParticles)$ to $(\targetDist_{\timeIndex-1},\normConst_{\timeIndex-1})$, normalizing flow $\baseFlow_\timeIndex$, unnormalized annealed targets $\unnormDensity_{\timeIndex-1}$ and $\unnormDensity_{\timeIndex}$ and resampling threshold $\resamplingThreshold\in\left[1/\numParticles,1\right)$.
		\STATE {\bf Output:} Particles at iteration $\timeIndex$: $\targetDist^\numParticles_{\timeIndex} = (\rvX_{\timeIndex}^{\particleIndex}, \particleWeightNorm_{\timeIndex}^\particleIndex)_{\particleIndex=1}^{\numParticles}$, approximation $\normConst^\numParticles_{\timeIndex}$ to $\normConst_{\timeIndex}$.%

			\STATE Transport particles: $\particleY^{\particleIndex}_{\timeIndex}=\baseFlow_\timeIndex(\particleX^{\particleIndex}_{\timeIndex-1})$.
			\STATE Compute IS weights: \\$\particleWeightUnnorm^{\particleIndex}_\timeIndex \leftarrow \particleWeightNorm_{\timeIndex-1}^\particleIndex \G_{\timeIndex}(\particleX_{\timeIndex-1}^\particleIndex)$ // unnormalized\\
				$\particleWeightNorm^{\particleIndex}_\timeIndex \leftarrow  \particleWeightUnnorm_{\timeIndex}^{\particleIndex}/\sum_{\particleIndexB=1}^\numParticles \particleWeightUnnorm_{\timeIndex}^{\particleIndexB}$ // normalized
			\STATE Estimate normalizing constant $\normConst_\timeIndex$: \\$\normConst_\timeIndex^\numParticles \leftarrow \normConst_{\timeIndex-1}^\numParticles \parens{\sum_{\particleIndex=1}^\numParticles \particleWeightUnnorm_\timeIndex^\particleIndex}$.
			\STATE Compute effective sample size $\textup{ESS}_\timeIndex^\numParticles$.
			\IF{$\textup{ESS}^{\numParticles}_{\timeIndex} \leq \numParticles\resamplingThreshold$} 
				\STATE Resample $\numParticles$ particles denoted abusively also $\particleY^\particleIndex_\timeIndex$ according to the weights $\particleWeightNorm_\timeIndex^\particleIndex$, then set $\particleWeightNorm_\timeIndex^\particleIndex = \frac{1}{\numParticles}$. 
			\ENDIF
			\STATE  Sample $ \particleX_{\timeIndex}^\particleIndex \sim \markovTransitionKernel_{\timeIndex}(\particleY^{\particleIndex}_{\timeIndex},\cdot)$. // MCMC
			\STATE Return $(\targetDist^\numParticles_{\timeIndex},\normConst_{\timeIndex}^\numParticles)$.
	\end{algorithmic}
\end{algorithm}

\begin{algorithm}[H]
\caption{CRAFT-training}\label{algo:craftTrain}
	\begin{algorithmic}[1]
		\STATE \textbf{Input:} Initial NFs $\{\baseFlow_\timeIndex\}_{1:\numParticles}$, number of particles $\numParticles$, unnormalized annealed targets $\{\unnormDensity_\timeIndex\}_{\timeIndex=0}^\numTransitions$ with $\unnormDensity_0{=}\targetDist_0$ and $\unnormDensity_\numTransitions{=}\unnormDensity$, resampling threshold $\resamplingThreshold\in\left[1/\numParticles,1\right)$.
		\STATE {\bf Output:} Learned flows $\baseFlow_\timeIndex$ and length $J$ sequence of approximations $(\targetDist^N_K,\normConst_K^N)$ to $(\targetDist_K,\normConst_K)$.%
		\FOR{$\craftIterIndex=1,\dots, \numCraftIters$}
		\STATE Sample $\particleX^\particleIndex_0 {\sim} \targetDist_0$ and set $\particleWeightNorm_0^\particleIndex {=} \frac{1}{\numParticles}$ and  $\normConst_0^\numParticles{=}1$.
		\FOR{$\timeIndex=1,\dots, \numTransitions$}
		    \STATE $\hat{h} \leftarrow \verb+flow-grad+\left(\baseFlow_\timeIndex, \targetDist_{\timeIndex-1}^\numParticles \right)$ using eqn \eqref{eq:gradientApproximation}.
			\STATE $\left(\targetDist^\numParticles_\timeIndex,\normConst_\timeIndex^\numParticles \right) {\leftarrow} \verb+ SMC-NF-step+\left(\targetDist^\numParticles_{\timeIndex-1},\normConst_{\timeIndex-1}^\numParticles, \baseFlow_\timeIndex\right)$
			\STATE Update the flow $\baseFlow_\timeIndex$ using gradient $\hat{h}$. 
		\ENDFOR
		\STATE Yield 	 $(\targetDist^\numParticles_{\numTransitions},\normConst_{\numTransitions}^\numParticles)$ and continue for loop.
	\ENDFOR
 \STATE Return learned flows $\{\baseFlow_\timeIndex\}_{k=1}^{\numTransitions}$.
	\end{algorithmic}
\end{algorithm}

\begin{algorithm}[H]
\caption{CRAFT-deployment}\label{algo:smcFlow}
	\begin{algorithmic}[1]
		\STATE \textbf{Input:} Fixed/trained NFs $\{\baseFlow_\timeIndex\}_{k=1}^{\numTransitions}$, number of particles $N$, unnormalized annealed targets $\{\unnormDensity_\timeIndex\}_{\timeIndex=0}^\numTransitions$ with $\unnormDensity_0{=}\targetDist_0$, resampling threshold $\resamplingThreshold\in\left[1/\numTransitions,1\right)$.
		\STATE {\bf Output:} Approximations $(\targetDist^\numParticles_\numTransitions,\normConst_\numTransitions^\numParticles)$ to $(\targetDist,\normConst)$.%
	    \vspace{.5\baselineskip}
		\STATE Sample $\particleX^\particleIndex_0 \sim \targetDist_0$ and set $\particleWeightNorm_0^i = \frac{1}{\numParticles}$ and  $\normConst_0^\numParticles=1$.
		\FOR{$\timeIndex=1,\dots, \numTransitions$}
			\STATE $\left(\targetDist^\numParticles_{\timeIndex},\normConst_{\timeIndex}^\numParticles \right){\leftarrow} \verb+ SMC-NF-step+\left(\targetDist^\numParticles_{\timeIndex-1},\normConst_{\timeIndex-1}^\numParticles, \baseFlow_\timeIndex\right)$
		\ENDFOR
		\STATE Return $(\targetDist^\numParticles_{\numTransitions},\normConst_{\numTransitions}^\numParticles)$.
	\end{algorithmic}
\end{algorithm}

\subsection{Limitations of ELBO based training}\label{section:conventional_elbo}
The idea of learning parameters of a forward generating Markov process $\forwardGenerating$ to approximate a target distribution $\pi_\numTransitions$ and its normalizing constant $\normConst_\numTransitions$ has already been explored in the literature, in particular in the context of VAEs training \citep{salimans2015markov,wunoe2020stochastic,Geffner:2021,Thin:2021,Zhang2021}. In these references, the standard approach consisting of minimizing the $\KL$ 
divergence between the distribution $\forwardGenerating$ of the forward generating process and a suitable augmented target $\augmentedTarget$ admitting $\targetDist_\numTransitions$ as marginal at time $\numTransitions$ is adopted:
\begin{equation}\label{eq:ELBO}
\KL[ \forwardGenerating || \augmentedTarget ] = \log \normConst_\numTransitions - \expect_{\forwardGenerating}\left[ \log \particleWeightUnnorm_{\numTransitions} \right].
\end{equation}

However, minimizing this Kullback--Leibler divergence, equivalently maximizing the ELBO objective $\expect_{\forwardGenerating}\left[ \log \particleWeightUnnorm_{\numTransitions} \right]$, requires differentiating through the forward sampler which can be challenging when MCMC kernels are included as we now discuss in more detail.  

{\bf Discontinuity of the forward sampler.} The most widely used MCMC transition kernels, such as Hamiltonian Monte Carlo (HMC) or Metropolis-adjusted Langevin algorithm (MALA), rely on an Metropolis accept/reject mechanism to ensure invariance. However, these mechanisms are discontinuous functions of the input random variables of the forward sampler. To avoid this issue and enable use of the reparametrization trick, a popular solution is to use approximate MCMC kernels without Metropolis correction which have continuous densities and then use an approximate reverse distribution \citep{salimans2015markov,wunoe2020stochastic,Geffner:2021,Thin:2021,Zhang2021}. The downside of this is the bias that accrues from not having the correction and the difficulty of approximating the reversal. This can mean many slow mixing proposals are required. 

It is desirable to use in the forward generating process a combination of standard Metropolis-corrected MCMC samplers (e.g. HMC, MALA) and NFs as used by CRAFT, Stochastic Normalizing Flows (SNFs) \cite{wunoe2020stochastic} and AFT \cite{Arbel2021}. In this case, the proposition below proven in Appendix \ref{section:elboGradient} and extending the results of \cite{Thin:2021} shows that the gradient of the ELBO in the flow parameters is the sum of two terms, one of which is a high variance score term. 
\begin{prop}
	Let $U$ be the set of continuous r.v.s (used to sample proposals) of distribution independent of any parameter and $A$ the set of discrete r.v.s (used for accept/reject steps) during the forward generating process. Let $p$ denote the joint distribution of $A,U$. The gradient of the ELBO $\expect_{\forwardGenerating}\left[ \log \particleWeightUnnorm_{\numTransitions} \right]$ in the flow parameters is:
\begin{align}\label{eq:gradientELBO}
		\expect_{p}\left[ \nabla \log (\particleWeightUnnorm_{\numTransitions}\circ \Phi) + \log(\particleWeightUnnorm_{\numTransitions}\circ \Phi) \nabla \log p  \right],	
\end{align}
	where $\Phi$ is a differentiable re-parameterization of the trajectory $X_{0:K}$ in terms of $U$ for any $A$, i.e. $X_{0:K}{=} \Phi(U,A)$. In particular, when the learned flows $(\baseFlow_\timeIndex)_{\timeIndex=1}^\numTransitions$ are optimal, i.e. $T^{\#}_{\timeIndex} \targetDist_{\timeIndex-1}=\targetDist_{\timeIndex}$, it holds that:
	\begin{align*}
	    	\expect_{p}\left[ \nabla \log (\particleWeightUnnorm_{\numTransitions}\circ \Phi) \right] = 
	    	\expect_{p}\left[ \log (\particleWeightUnnorm_{\numTransitions}\circ \Phi) \nabla \log p  \right] =0.	
	\end{align*}
\end{prop}
The first term on the r.h.s. of the expression \eqref{eq:gradientELBO} corresponds to a reparameterization trick term. The second term, however, requires computing a high variance score $\nabla\log p$. 
As pointed out by \citet{Thin:2021}, 
algorithms such as SNFs with Metropolis-adjusted MCMC kernels are omitting the second score term when optimizing the ELBO.
While this term happens to vanish in the particular case where the flows are optimal, its contribution can, in general, be important especially when initializing the algorithm  with sub-optimal flows. 

{\bf Discontinuity of the resampling steps.}
The literature has mostly focused on scenarios where the forward generating process $\forwardGenerating$ is a simple Markov process. SMC forward generating processes as used in CRAFT could be exploited to obtain a tighter ELBO and have been used in the context of state-space models \cite{maddison2017filtering,le2017auto,naesseth2017variational}. However, the resampling steps of SMC correspond to sampling discrete distributions and lead to high variance gradient estimates. Omitting them can introduce significant bias \cite{corenflos2021differentiable}. These difficulties are further exacerbated in our context since, as discussed earlier, the MCMC steps are not differentiable either. 

{\bf Emergence of annealing requires mixing.} In \citep{wunoe2020stochastic} the ELBO corresponding to equation \eqref{eq:ELBO} has the form of the standard VI with NFs objective, with additional weight terms entering into $\particleWeightUnnorm_{\numTransitions}$ due to the MCMC steps and annealing schedule. The additional log-weight corrections are    
\begin{equation}\label{eq:updateweightSNF}
\Delta \log \particleWeightUnnorm^{\text{SNF MC}}_{\timeIndex}= \log \unnormDensity_{\timeIndex}(\x^{\text{SNF}}_{\timeIndex}) - \log \unnormDensity_{\timeIndex}(\hat{\x}^{\text{SNF}}_{\timeIndex}), \end{equation}   
\noindent where $\hat{\x}^{\text{SNF}}_{\timeIndex}$ is the value of the sample after the MCMC at temperature $\timeIndex$ has been applied and $\x^{\text{SNF}}_{\timeIndex}$ is the value of the sample before it (see Appendix \ref{sec:appextendedproposaltargetSNF}). This is the only way the annealed densities enter into the Stochastic Normalizing Flow ELBO. Since this additional term is zero in the limit where the Markov kernel does not mix, it follows that the value of the SNF ELBO reduces to the VI with normalizing flows objective in this case. Consequently, for the training objective to be significantly different from the one without MCMC and annealing, this method requires the MCMC kernels to mix sufficiently, which may not be the case in challenging examples.

\subsection{CRAFT training}\label{section:craft_training}

Here we describe a method that neatly sidesteps the issues described with the standard ELBO. We take the learning objective to have the following form:
\begin{align}\label{eq:overallObjective}
\overallObjective &= \sum_{\timeIndex=1}^{\numTransitions} \KL[T^{\#}_{\timeIndex} \targetDist_{\timeIndex-1} || \targetDist_{\timeIndex}]
\end{align}
\noindent where the sum runs over each transition between temperatures. The minimum of the objective is zero and this is attained when each flow transports perfectly between successive temperatures. In general, breaking the annealing task down term-by-term means that we can reduce the mode seeking effect of the reverse $\KL$ divergence. Each $\KL$ term in the sum can be written as follows:
\begin{subequations}
	\begin{gather}
	\KL[\baseFlow^{\#}_{\timeIndex} \targetDist_{\timeIndex-1} || \targetDist_{\timeIndex}] = \expect_{ \targetDist_{\timeIndex-1}}\left[ \deltaDensity_{\timeIndex}\right] + \log \left(\frac{\normConst_\timeIndex}{\normConst_{\timeIndex-1}}\right),\label{eq:klExpand}\\
		\deltaDensity_{\timeIndex}(x) := \log \frac{\unnormDensity_{\timeIndex-1}(x)}{ \unnormDensity_{\timeIndex}(\baseFlow_{\timeIndex}(x))} - \log|\nabla_{x} \baseFlow_{\timeIndex}(x)|.\label{eq:D_l}
	\end{gather}
\end{subequations}
The expectation over $\targetDist_{\timeIndex-1}$ on the RHS of the equation is approximated using a streaming SMC approximation of $\targetDist_{\timeIndex-1}$
 which will always be based on the corresponding learnt sampler for the current optimization step: 
 \begin{equation}\label{eq:lHat}
 \KL[\baseFlow^{\#}_{\timeIndex} \targetDist_{\timeIndex-1} || \targetDist_{\timeIndex}] 
 - \log \left(\frac{\normConst_\timeIndex}{\normConst_{\timeIndex-1}}\right) \approx \sum_{\particleIndex} \particleWeightNorm_{\timeIndex-1}^{\particleIndex} \deltaDensity_{\timeIndex}(\particleX_{\timeIndex-1}^{\particleIndex}).
 \end{equation}
 The terms inside the summation require that we can evaluate the output of the normalizing flow $T_{\timeIndex}(X)$ and the log determinant of the Jacobian $\log|\nabla_{x} \baseFlow_{\timeIndex}(x)|$. Similarly the gradients with respect to flow parameters $\genericFlowParams_{\timeIndex}$ of flow $\baseFlow_{\timeIndex}$ are approximated:
 
\begin{equation}\label{eq:gradientApproximation}
\frac{\partial \overallObjective}{\partial \genericFlowParams_{\timeIndex}} \approx \sum_{\particleIndex} \particleWeightNorm_{\timeIndex-1}^{\particleIndex} \frac{\partial \deltaDensity_{\timeIndex}(\particleX_{\timeIndex-1}^{\particleIndex})}{\partial \genericFlowParams_{\timeIndex}}.
\end{equation}
 When the flows transport perfectly, the gradient $\frac{\partial \overallObjective}{\partial \genericFlowParams_{\timeIndex}}$ will have true value zero and there will be no bias in this particle approximation to it. More generally the gradient estimate will be biased since it is based on a normalized importance estimator.
 We will analyze this aspect further in Section \ref{section:particleKL}, but for now we note that the normal guarantees from the SMC literature imply the gradient is consistent in the number of particles and the variance and bias are of order $O(1/\numParticles)$ \citep{Del-Moral:2013a}.

The proposed algorithm is described in Algorithm \ref{algo:craftTrain}. Each training loop iteration of CRAFT looks like a iteration of the test time sampler in Algorithm \ref{algo:smcFlow} with gradient estimation and flow parameter updates added. Note that the parameter update is applied after the flow transport step for that temperature so the first effect it has will be on the particles that come through in the next step. This allows us to view CRAFT as an optimization of the objective \eqref{eq:overallObjective} and will allow stability analysis that we will detail later in the text. It also means that the normalizing constant estimators produced after each step are unbiased. Since all the constants produced in this way depend on the optimization sequence they are not independent. At test time when the parameters are fixed the independence is restored.

{\bf Relationship to AFT.} We now describe the relationship between CRAFT and the AFT method of \citep{Arbel2021}. Relating them is made more challenging by the fact there are two variants of the AFT algorithm described in \citep{Arbel2021}. For clarity we include both in our supplement as Algorithms \ref{algo:AFTsimple} and \ref{algo:AFTpractical}. We call Algorithm  \ref{algo:AFTsimple} \emph{simple AFT}- it is easier to describe and analyse but is not actually used in practice by \citep{Arbel2021} for reasons we will describe. Instead they use what we call \emph{practical AFT} (Algorithm \ref{algo:AFTpractical}) a more complicated version that performs better in practice.  Relative to CRAFT, both variants of AFT may be viewed as a one pass greedy optimization of the objective \eqref{eq:overallObjective}. The closest point of algorithmic comparison to CRAFT is simple AFT- we show the line difference to CRAFT in Algorithm \ref{algo:CRAFTdiff}. At each temperature step simple AFT optimizes the particle approximated loss to its numerical minimum, before then updating the particles using the new flow.  

We now describe the \emph{sample replenishment problem} in AFT. The learning of the flows is hampered by the fact that the sample complexity of estimation can be higher than the finite number of available particles. In simple AFT this manifests as over-fitting of the flows to the available particles and large biases in expectations. Regenerating new samples at each temperature separately would require computation quadratic in the number of temperatures, which is too slow. To mitigate the over-fitting in a manor consistent with the one pass paradigm of the paper, the authors added validation samples which were used for early stopping. Test samples were also added to remove finite sample residual bias. This significantly increased the complexity of the AFT method, changing simple AFT (Algorithm \ref{algo:AFTsimple}) to practical AFT (Algorithm \ref{algo:AFTpractical}). This gave substantial empirical improvements but the fundamental limitation of the sample replenishment problem remains, arising from the desire to use one temperature pass. Even when the particles are a perfect approximation to the desired distribution at each temperature, if the sample complexity of estimating the training gradients of the flows is high relative to the number of particles then the practical AFT method still struggles to train.

By having multiple independent passes instead of one, CRAFT is able to fully address the sample replenishment problem in a different way. Since in CRAFT the parameter update is applied after flow transport at each temperature the analysis of the algorithm is simpler than that required for AFT, where sophisticated arguments must be invoked to ensure consistency of the particles. Later in Section \ref{section:particleKL} we will see that we can analyse CRAFT from the perspective of unbiased gradients. This interpretation does not apply to either variant of AFT. 

From a practical perspective CRAFT performs better than AFT (Section \ref{section:CRAFTvsAFT}). This is significant because \citep{Arbel2021} already showed they could train samplers that performed well relative to strong baselines like SMC. CRAFT is easier to tune and more robust to hyperparameter misspecification than AFT. Conceptually it is much simpler, to the extent that it is easier to describe CRAFT as a standalone algorithm than in terms of the AFT algorithm on which it builds. CRAFT is closer to a traditional machine learning training paradigm with gradients applied to a single objective. This makes it easier to adopt and to scale through parallelism.  

Recent work from \citep{Zimmermann:2021} investigates a framework that reduces to the AFT objective \eqref{eq:overallObjective} for deterministic forward and backward transitions. Inspired by \citep{Arbel2021} they investigate some toy examples with normalizing flows. There are good reasons to concentrate on the normalizing flow case as we have done. The reverse distribution is analytic and optimal for a flow but can be difficult to approximate otherwise \citep{Thin:2021}. Further \citet{Zimmermann:2021} do not use Markov transition kernels in practice, which we found to be essential for good performance. 

\subsection{Using CRAFT within Particle MCMC}
Whilst the SMC-NF estimator of the normalizing constant is unbiased, in general estimates of the expectations w.r.t. the target are asymptotically consistent but exhibit a $O(1/\numParticles)$ bias. In situations requiring high accuracy expectations it is desirable to have a method returning asymptotically unbiased estimates.  However, increasing $\numParticles$ for the bias to be negligible will often not be feasible for challenging applications because of memory requirements. MCMC algorithms have the advantage of providing consistent estimate of these expectations as compute time increases without having to store an increasing number of samples. Particle MCMC methods \cite{Andrieu2011} provide a way to bring such benefits to SMC samplers. In particular, the so-called Particle Independent Metropolis--Hastings is an MCMC sampler that uses an SMC sampler with $\numParticles$ particles and provides consistent estimates of expectations for any $\numParticles$ as the number of iterations increases; see Appendix \ref{sec:particleMCMCdescription} for details. Since a single pass of CRAFT with fixed parameters can be thought of as an SMC sampler with additional deterministic transformations it can be used here. Computational trade-offs for Particle MCMC are analyzed by \citet{pitt2012some}.%

\section{Analysis of the training objective}
\subsection{Reformulating the CRAFT objective}
The CRAFT objective \eqref{eq:overallObjective} can be rewritten as a single $\KL$ divergence between certain product distributions:
\begin{equation}\label{eq:CRAFTobjective}
\overallObjective = \KL\left[ \prod_{\timeIndex=1}^{\numTransitions} \baseFlow^{\#}_{\timeIndex} \targetDist_{\timeIndex-1} || \prod_{\timeIndex=1}^{\numTransitions} \targetDist_{\timeIndex} \right].
\end{equation}

We have adopted the notation that the product symbol $\prod$ can be used to denote multi-variable product measures. The proof above effectively applies the multi-variable generalization of the result for the $KL$-divergence between two product measures, $\KL[U{\times} V || F {\times} G] {=} \KL[U || F] {+} \KL[V || G]$. ~Despite the non-standard form of the $\KL$ divergence, it is also possible to obtain a bound on the normalizing constant
\begin{equation}
\log \normConst_\numTransitions \geq
-\sum_{\timeIndex=1}^{\numTransitions}\expect_{X \sim \targetDist_{\timeIndex-1}}\left[ \deltaDensity_\timeIndex \right]
\end{equation}
This bound can be obtained starting from $\overallObjective \geq 0$, expanding each term using \eqref{eq:klExpand} and noting that there is a telescoping cancellation between the intermediate normalizing constants. The bound is distinct from the one obtained if we take the logarithm of the unbiased estimate used in SMC samplers. In practice the bound is estimated using the particle estimate for each temperature and is effectively the training objective of CRAFT. %

\subsection{Re-thinking the KL criterion at the particle level}\label{section:particleKL}
We discuss here a re-interpretation of the CRAFT training procedure in terms of KL-divergences between particle approximations and targets. An advantage of this interpretation is that it shows that we compute an unbiased gradient of a modified objective that is well-motivated even when the number of particles is moderate. 

CRAFT training requires minimizing the objective \eqref{eq:overallObjective}. As we do not have access to $\targetDist_{\timeIndex-1}$, CRAFT approximates the intractable gradient of this objective using equation \eqref{eq:gradientApproximation}. We will compactly refer to the particle approximation as a weighted sum of delta masses $\targetDist^\numParticles_{\timeIndex-1}(\mathrm{d}\x){=}\sum_{\particleIndex=1}^\numParticles \particleWeightNorm^\particleIndex_{\timeIndex-1} \delta_{\particleX^\particleIndex_{\timeIndex-1}}(\mathrm{d}\x)$. While the use of such an approach could be of concern when $\numParticles$ is moderate or at initialization, the following proposition shows that such concerns are ill-founded and that this approximate SMC based gradient is also the \emph{unbiased} gradient of a well-defined and intuitive objective.
\begin{prop}\label{prop:particleobjectiveCRAFT}
Let $\hat{\targetDist}^\numParticles_{\timeIndex-1}(\mathrm{d}\x)=\mathbb{E}[\targetDist^\numParticles_{\timeIndex-1} (\mathrm{d}\x)]$ denote the expectation of the random SMC approximation $\targetDist^\numParticles_{\timeIndex}$ of $\targetDist_{\timeIndex-1}$  w.r.t. to the law of the SMC-NF algorithm. An unbiased gradient of the objective 
$\KL[\baseFlow_{\timeIndex}^{\pushForward}\hat{\targetDist}^\numParticles_{\timeIndex-1} || \targetDist_{\timeIndex} ]$ is given by 
\begin{equation}\label{eq:unbiasedgradient}
    \mathbb{E}_{X\sim \targetDist^\numParticles_{\timeIndex-1}}[\nabla_{\genericFlowParams_{\timeIndex}} \deltaDensity_{\timeIndex}(\particleX)].
\end{equation}
\end{prop}
The l.h.s of this KL is the pushforward of the ``average'' SMC approximation to $\targetDist_{\timeIndex-1}$ through the flow  $\baseFlow_{\timeIndex}$. 

To help understand this result and how it is derived, consider a simpler scenario where the random approximation $\targetDist^\numParticles_{\timeIndex-1}$ of $\targetDist_{\timeIndex-1}$ has been obtained by a batch parallel importance sampling method. That is we sample $\particleX_{\timeIndex-1}^\particleIndex {\overset{i.i.d.}{\sim}} q_{\timeIndex-1}$ for $\particleIndex{=}1,\dots,\numParticles$, compute  $\particleWeightUnnorm(X_{\timeIndex-1}^{\particleIndex}){=}\targetDist_{\timeIndex-1}(X_{\timeIndex-1}^\particleIndex)/q_{\timeIndex-1}(X_{\timeIndex-1}^\particleIndex)$. We then define $\particleWeightNorm^\particleIndex_{\timeIndex-1} {\propto} \particleWeightUnnorm(X_{\timeIndex-1}^\particleIndex)$ such that $\sum_{\particleIndex=1}^N \particleWeightNorm^\particleIndex_{\timeIndex-1}{=}1$ and finally return $\targetDist^\numParticles_{\timeIndex-1}(\mathrm{d}\x)$. If we average over the random particle locations $\particleX^{1:\numParticles}_{\timeIndex-1}$, we obtain
\begin{equation}\label{eq:averagerandommeasure}
\hat{\targetDist}^\numParticles_{\timeIndex-1}(\mathrm{d}\x)=\mathbb{E}_{\particleX^\particleIndex_{\timeIndex-1} \overset{\textup{i.i.d.}}{\sim} q_{\timeIndex-1}}[\targetDist^\numParticles_{\timeIndex-1} (\mathrm{d}\x)].
\end{equation}
Contrary to $\targetDist^\numParticles_{\timeIndex-1} (\mathrm{d}\x)$ which is a discrete measure,
the distribution 
$\hat{\targetDist}^\numParticles_{\timeIndex-1}(\mathrm{d}\x)$ admits a density and its Kullback--Leibler divergence with $\klTargetDist=(\baseFlow_{\timeIndex}^{-1})^{\pushForward}\targetDist_{\timeIndex}$ is well-defined. 
It then follows directly from \eqref{eq:averagerandommeasure}, diffeomorphism invariance, and iterated expectations that 
\begin{align}
    \KL[\baseFlow_{\timeIndex}^{\pushForward}\hat{\targetDist}^\numParticles_{\timeIndex-1} &|| \targetDist_{\timeIndex} ]
    =\mathbb{E}_{\particleX \sim \hat{\targetDist}^\numParticles_{\timeIndex-1}}\Big[\log \frac{ \hat{\targetDist}^\numParticles_{\timeIndex-1}(X)}{ \klTargetDist(X)}\Big] \\
    &=\mathbb{E}_{\particleX^\particleIndex_{\timeIndex-1} \overset{\textup{i.i.d.}}{\sim} q_{\timeIndex-1}}\Big[\mathbb{E}_ {\particleX \sim \targetDist^\numParticles_{\timeIndex-1}}\Big[\log \frac{\hat{\targetDist}_{\timeIndex-1}(X)}{ \klTargetDist(X)}\Big]\Big] \nonumber.
\end{align}
A direct consequence of this identity is that an unbiased gradient of $\KL[\baseFlow_{\timeIndex}^{\pushForward}\hat{\targetDist}^\numParticles_{\timeIndex-1} || \targetDist_{\timeIndex} ]$ is indeed given by \eqref{eq:unbiasedgradient}.
To prove Proposition \ref{prop:particleobjectiveCRAFT}, we extend this argument to the scenario where   $\hat{\targetDist}^\numParticles_{\timeIndex-1}(\mathrm{d}\x)$ has been obtained by using an SMC sampler where the particles are not independent because of resampling;  see Appendix \ref{sec:appparticleintepretationCRAFT}.

Taking a step back and returning to the original objective we note that once we start considering the marginal particle distribution $ \hat{\targetDist}^\numParticles_{\timeIndex-1}$ instead of the target distribution $\targetDist_{\timeIndex-1}$ the flow parameters $\genericFlowParams$ effect the $\KL$ divergences that follow them, since later particle distributions depend on them. This additional effect is not included in the CRAFT updates. Each flow takes whatever particle distribution it receives from the steps and is trained only to reduce the $\KL$ of the push-forward to the next target distribution $\targetDist_{\timeIndex}$, without reference to what follows.

\section{Experiments}

In this section we empirically investigate the performance of CRAFT. First, in Section \ref{section:CRAFTvsAFT} we give a case study demonstrating the empirical benefit of CRAFT relative to AFT, then in Section \ref{section:CRAFTvsSNF} we show that CRAFT outperforms Stochastic Normalizing flows in two challenging examples. We then show a compelling example use case for CRAFT as a learnt proposal for a particle MCMC sampler applied to lattice field theory. Code for the algorithms and examples \ifarxiv can be found at \url{https://github.com/deepmind/annealed_flow_transport}.\hspace{3 pt}\else is included in the supplementary material.\fi Further experiments and details are included in the Appendix. 

\subsection{Empirically comparing CRAFT and AFT}\label{section:CRAFTvsAFT}

Our first point of comparison for CRAFT is with AFT \citep{Arbel2021}. We use what we call the practical AFT method in Section \ref{section:craft_training}. To illustrate the benefit of solving the AFT sample replenishment problem we show the effect of varying the batch size in both algorithms. Both methods are given the same total budget of particles at train and test time. To highlight that the sample replenishment problem cannot be mitigated simply by using more MCMC we gave AFT 100 times more MCMC updates than CRAFT at train and test time.  We use the 1024 dimensional log Gaussian Cox process (LGCP) example which is the most challenging from \citep{Arbel2021}. As in that paper, we used a diagonal affine flow.

\begin{figure}[ht]
\begin{center}
\centerline{\includegraphics[width=0.95\linewidth]{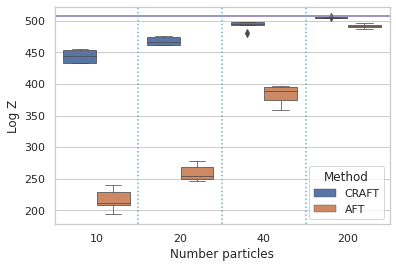}}
\caption{Comparison of normalizing constants between CRAFT and AFT for different numbers of particles. AFT is given 100 times more train/test MCMC steps than CRAFT. A gold standard value is shown in magenta.}\label{fig:AFTvsCRAFT}
\end{center}
\vskip -0.2in
\end{figure}

Figure \ref{fig:AFTvsCRAFT} shows the results. We see that CRAFT outperforms AFT for all number of particles considered. This can be understood in terms of our better solution to the sample replenishment problem as described in Section \ref{section:craft_training}. Since many of the most challenging modern sampling problems also have high memory usage this is also of considerable practical significance.

\subsection{Comparing CRAFT and SNFs}\label{section:CRAFTvsSNF}

We now compare CRAFT training with that of the corresponding SNFs. For the Markov transition kernel both methods use full Metropolis corrected HMC. We use the same normalizing flow family for both methods. As such this constitutes a direct comparison of the two approaches. We ran timed training experiments for both methods computing the corresponding unbiased estimates for the normalizing constant as it progressed. Note that at test time these estimates could be obtained faster by running the learnt sampler. We consider two different target densities following \citep{Arbel2021} namely the 30 dimensional variational autoencoder latent space and the LGCP example. For the VAE example we used an affine inverse autoregressive flow \citep{Kingma:2016}. For the LGCP we again used a diagonal affine flow.  

\begin{figure}[ht!]
\begin{center}
\centerline{\includegraphics[width=0.95\linewidth]{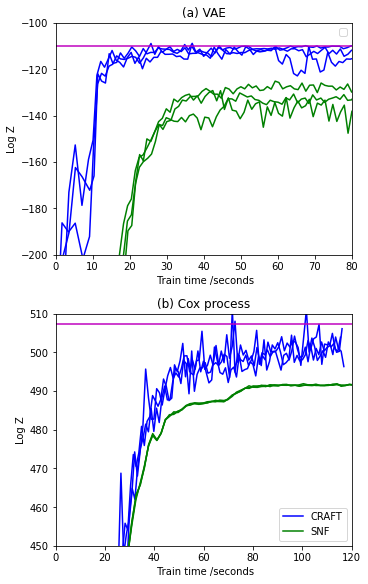}}
\caption{Timed comparison of normalizing constant estimates produced by CRAFT and SNF during training (higher is better). Three independent repeats are shown. A gold standard value is shown in magenta.}
\label{fig:SNFvsCRAFT}
\end{center}
\vskip -0.2in
\end{figure}

Figure \ref{fig:SNFvsCRAFT} shows the results. In both cases we find that CRAFT converges to a better value than SNFs. We attribute the worse final value for SNFs to the issues with the training discussed in Section \ref{section:conventional_elbo}.   

\subsection{CRAFT based Particle MCMC for lattice $\phi^4$ theory}\label{section:phiFour}

\begin{figure*}[ht!]
\vskip 0.2in
\begin{center}
\centerline{\includegraphics[width=0.92\linewidth]{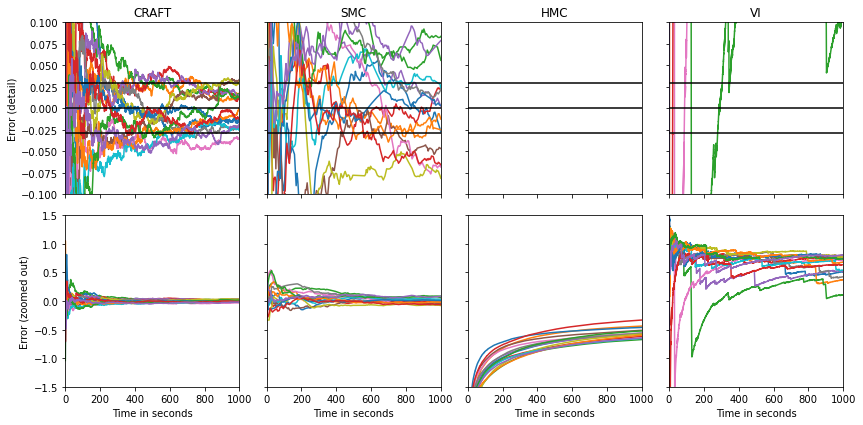}}
\caption{Timed comparison of MCMC methods for the $\phi^4$ example, based on fifteen repeats. CRAFT, SMC and VI serve as proposal mechanisms for Particle MCMC. HMC is applied directly to the target. Error is in estimating two point susceptibility, an example, physically relevant expectation. Note HMC never reaches the detailed level of error in the top row.}
\label{fig:phiFourComparison}
\end{center}
\vskip -0.2in
\end{figure*}

A classic proving ground for sampling algorithms is in the area of lattice field theory. Indeed, it was this application that motivated Hamiltonian/Hybrid Monte Carlo \citep{Duane1987}. State of the art quantum chromodynamics (QCD) calculations \citep{Borsanyi2021} use lattice sampling run on multiple supercomputers. There has been a recent surge of interest in using normalizing flows to improve sampling in such models starting with \citep{Albergo2019}, so it is natural to investigate the performance of CRAFT in this area.

Of the available lattice field theories we use the Euclidean $\phi^4$ theory in two dimensions following \citep{Albergo2019, Albergo2021}. We start from the continuum action:
\begin{equation*}
\action_{\text{cont}}[\phi] = \int \left\lbrace ||\nabla \phi(x) ||_2^{2} + \mass^2 \phi(x)^2 + \coupling \phi(x)^4 \right\rbrace \mathrm{d}x,
\end{equation*}
which we discretize on a $14 \times 14$ lattice, using lattice units, and periodic boundary conditions to obtain:
\begin{subequations}\label{eq:latticePhiFour}
\begin{align}
\action_{\text{latt}}(\phi) =& \sum_{\latticeX} \lbrace \phi(\latticeX) \discreteNeighbour(\latticeX) + \mass^2 \phi(\latticeX)^2 + \coupling \phi(\latticeX)^4 \rbrace, \\
\discreteNeighbour(\latticeX) :=& \sum_{\latticeDirection} \left[ 2 \phi(\latticeX) - \phi(\latticeX + \latticeUnitVector_{\latticeDirection}) - \phi(\latticeX - \latticeUnitVector_{\latticeDirection}) \right].
\end{align}
\end{subequations}
\noindent Here the sum over $\latticeX$ runs over all lattice sites. The summation $\latticeDirection \in \lbrace \latticeSizeX, \latticeSizeY \rbrace$ runs over the dimensions of the lattice and $\latticeUnitVector_{\latticeDirection}$ defines a single lattice step the in direction $\latticeDirection$. $\coupling \geq 0$ defines the \emph{coupling parameter} and $\mass^2$ is the \emph{mass squared} which counter-intuitively may be positive or negative. A probability distribution over the values of the field at the lattice sites is then defined using $p(\phi) = \frac{1}{\normConst}\exp\left\lbrace-\action_{\text{latt}}(\phi)\right\rbrace$. For positive $\coupling$ and large $\phi$ the quartic term will dominate giving highly non-Gaussian tails. For negative $\mass^2$, which will be our focus, the marginal distributions are bimodal.

As is typical of physical sciences applications, symmetry is crucial in this example. The target is invariant under any translation on the lattice. In Appendix \ref{section:symmetry} we show that the corresponding requirement on flows between successive temperatures is translation equivariance. We incorporate this in to the flow models under consideration using convolutional affine coupling layers \citep{Dinh2017} with periodic boundary conditions. 

Since high accuracy is required we are interested in methods that are asymptotically unbiased for general expectations. We take physically relevant expectations from \citep{Albergo2019} namely the `two point susceptibility' in the main text and the `Ising energy density' in the supplement -both have similar results. We use Particle MCMC with three different independent proposal types. These are SMC, VI with normalizing flows (and no annealing) and CRAFT. We also compare against HMC 
that directly targets the target density.   
The results are shown in Figure \ref{fig:phiFourComparison}. We see that HMC struggles with this target density because it is unable to mix between modes. Despite producing samples quickly, the VI proposal fails to cover all of the modes because of the use of reverse KL divergence. The SMC proposal does much better. As an annealing method it is less susceptible to multi-modality. Finally CRAFT converges the fastest, improving on SMC without flows and avoiding the mode seeking behaviour of pure VI. Note that by scaling arguments if we accelerate an MCMC method by a factor of $R$ we expect that errors in expectations will be reduced by a factor of $\sqrt{R}$ on such error plots.

\section{Conclusion}

In this paper we have proposed and analyzed the CRAFT algorithm. At test time it is an SMC sampler with added flow transport steps. The training algorithm uses particle estimation to optimize a training objective that explicitly promotes transport between annealing temperatures. We have given a thorough analysis of the training including an insightful particle interpretation. We have shown that CRAFT compares well empirically and conceptually with existing methods to estimate flows in this context. 

In terms of where such efforts fit into the bigger picture, there are many applications where high accuracy expectations are required and algorithms like Particle MCMC become necessary due to the challenging nature of the target. As such we believe the speed up we have achieved in the quantum field theory application is exciting both for that field and others.

\ifarxiv
\section{Acknowledgements}

The authors wish to thank S\'{e}bastien Racani\`{e}re, Michalis Titsias, Phiala Shanahan, Jonas K\"{o}hler, Andy Ballard, Laurence Midgley and the anonymous reviewers.  
\fi

\bibliographystyle{plainnat}  
\bibliography{references,biblio}

\clearpage
\onecolumn
\appendix
\icmltitle{Continual Repeated Annealed Flow Transport Monte Carlo: Supplementary Material}
\section{Extended proposal and target distributions}\label{sec:appextendedproposaltargetSNF}
Assume the transport maps $T_l$ are fixed. We detail the extended proposal and target distributions used by Stochastic Normalizing Flows \cite{wunoe2020stochastic}. These were introduced early on in \cite{vaikuntanathan2011escorted}. The same target and proposal are also used by CRAFT and Annealed Flow Transport  \cite{Arbel2021} and the Gibbs flow algorithm \cite{heng2015gibbs} when resampling is omitted. We include here these distributions for sake of completeness.

We sample $X_0\sim \pi_0(\cdot)$ at $k=0$ then use $Y_k=T_k(X_{k-1})$ followed by $X_k \sim \mathcal{K}_k(Y_k,\cdot)$ at time $k\geq 1$. To simplify presentation, we do not use measure-theoretic notation. Hence, using the notation  $M^{\textup{trans}}_{l}(x,x')=\delta_{T_l(x)}(x')$ and $M^{\textup{mut}}_{l}(x,x')=\mathcal{K}_l(x,x')$, we obtain the following proposal at time $k$ after the transport step
\begin{equation}\label{eq:extendedproposalnoresampling}
\bar{\eta}_k(x_{0:k-1},y_{1:k})=\pi_0(x_0) \left(\prod_{l=1}^{k-1} M^{\textup{trans}}_{l}(x_{l-1},y_l) M^{\textup{mut}}_{l}(y_l,x_l)\right)  M^{\textup{trans}}_{k}(x_{k-1},y_k) ,
\end{equation}
and the target is
\begin{equation}\label{eq:extendedtargetnoresampling}
\bar{\pi}_k(x_{0:k-1},y_{1:k})=\pi_k(y_k) L^{\textup{trans}}_{k-1}(y_k,x_{k-1})\left(\prod_{l=1}^{k-1} L^{\textup{mut}}_{l-1}(x_{l},y_{l}) L^{\textup{trans}}_{l-1}(y_{l},x_{l-1})\right),
\end{equation}
where $L^{\textup{trans}}_{l-1}(x,x')=\delta_{T_l^{-1}(x)}(x')$ and $L^{\textup{mul}}_{l-1}(x,x')=\pi_l(x')M^{\textup{mul}}_l(x',x)/\pi_l(x)$.
After the mutation step at time $k$, the proposal is
\begin{equation}
\bar{\eta}_k(x_{0:k},y_{1:k})=\bar{\eta}_k(x_{0:k-1},y_{1:k}) M^{\textup{mut}}_{k}(y_k,x_k)= \pi_0(x_0)\left(\prod_{l=1}^k M^{\textup{trans}}_{l}(x_{l-1},y_l) M^{\textup{mut}}_{l}(y_l,x_l)\right)
\end{equation}
and the target is 
\begin{equation}
\bar{\pi}_k(x_{0:k},y_{1:k})=\pi_k(x_k)\left(\prod_{l=0}^{k-1} L^{\textup{mut}}_{l}(x_{l+1},y_{l+1}) L^{\textup{trans}}_{l}(y_{l+1},x_l)\right).
\end{equation}
Hence the incremental weight after a transport term at time $k$ is of the form 
\begin{equation}
\frac{\bar{\pi}_k(x_{0:k-1},y_{1:k})}{\bar{\eta}_k(x_{0:k-1},y_{1:k})}=\frac{\bar{\pi}_{k-1}(x_{0:k-1},y_{1:k-1})}{\bar{\eta}_{k-1}(x_{0:k-1},y_{1:k-1})}\underbrace{\frac{\pi_k(y_k)L^{\textup{trans}}_{k-1}(y_k,x_{k-1})}{\pi_{k-1}(y_{k-1})M^{\textup{trans}}_{k}(x_{k-1},y_k)}}_{\textup{incremental weight}=\frac{Z_{k-1}}{Z_k}G_{k,T_k}(x_{k-1})},
\end{equation}
while after the mutation step it is of the form
\begin{equation}
\frac{\bar{\pi}_k(x_{0:k},y_{1:k})}{\bar{\eta}_k(x_{0:k},y_{1:k})}=\frac{\bar{\pi}_k(x_{0:k-1},y_{1:k})}{\bar{\eta}_k(x_{0:k-1},y_{1:k})}\underbrace{\frac{\pi_k(x_k)L^{\textup{mul}}_{k-1}(x_k,y_k)}{\pi_{k}(y_k)M^{\textup{mul}}_{k}(y_k,x_k)}}_{\textup{incremental weight}=1}.
\end{equation}

As shown in Appendix B.2 of \cite{Arbel2021}, we can also integrate out the random variables $Y_{1:k}$ in these expressions as we can collapse the transport step and mutation step into one single Markov kernel
\begin{align}\label{eq:collapsedkernel}
M_l^{\textup{col}}(x_{l-1},x_l) &=  \int M^{\textup{trans}}_{l}(x_{l-1},y_l) M^{\textup{mut}}_{l}(y_l,x_l)\mathrm{d}y_l \nonumber\\
                &= \int \delta_{T_l(x_{l-1})}(y_l)\mathcal{K}_l(y_l,x_l)\mathrm{d}y_l \nonumber\\
                &= \mathcal{K}_l(T_l(x_{l-1}),x_l).
\end{align}
Similarly we can collapse the backward kernels that have been used to defined the extended target distributions $\bar{\pi}_k$
\begin{align}\label{eq:collapsedbackwardkernel}
L_{l-1}^{\textup{col}}(x_l,x_{l-1}) &=\int  L^{\textup{mut}}_{l-1}(x_{l},y_{l}) L^{\textup{trans}}_{l-1}(y_{l},x_{l-1}) \mathrm{d}y_l \nonumber\\
&=\frac{\pi_l(T_l(x_{l-1})) |\nabla T_l(x_{l-1})| \mathcal{K}_l(T_l(x_{l-1}),x_l)}{\pi_l(x_l)},
\end{align}

so the collapsed proposal and target can be written as \begin{equation}\label{eq:extendedproposalcollapsednoresampling}
\bar{\eta}_k(x_{0:k})=\pi_0(x_0) \prod_{l=1}^k M^{\textup{col}}_{l}(x_{l-1},x_l),
\end{equation}
\begin{equation}\label{eq:extendedtargetcollapsednoresampling}
\bar{\pi}_k(x_{0:k})=\pi_k(x_k)\prod_{l=0}^{k-1} L^{\textup{col}}_{l}(x_{l+1},x_{l}).
\end{equation} 

For $k=K$, we write $\bar{\eta}_K=\bar{\eta}$ and $\bar{\pi}_K=\bar{\pi}$.

It follows that 
\begin{align}
    \frac{Z_K \bar{\eta}(x_{0:K})}{\bar{\pi}(x_{0:K})}:=w_K(x_{0:K-1})&=\frac{ \gamma_{K}(x_K)}{\gamma_0(x_0)}\prod_{l=1}^{K}\frac{L^{\textup{col}}_{k-1}(x_{k}, x_{k-1})}{ M^{\textup{col}}_k(x_{k-1}, x_k)} \label{eq:decompoweightSNF}\\
&=\frac{ \gamma_{K}(x_K)\gamma_{K-1}(x_{K-1})\cdots \gamma_1(x_1)}{\gamma_0(x_0)\gamma_{1}(x_{1})\cdots \gamma_{K-1}(x_{K-1})}\prod_{k=1}^{K}\frac{L^{\textup{col}}_{k-1}(x_{k}, x_{k-1})}{ M^{\textup{col}}_k(x_{k-1}, x_k)}\nonumber\\
&=\prod_{k=1}^{K}\frac{ \gamma_{k}(x_k) L^{\textup{col}}_{k-1}(x_{k}, x_{k-1})}{  \gamma_{k-1}(x_{k-1})M^{\textup{col}}_k(x_{k-1}, x_k)} \nonumber\\
&=\prod_{l=1}^K G_k(x_{k-1})\label{eq:productincrementalweight},
\end{align} 
where, from (\ref{eq:collapsedkernel}) and (\ref{eq:collapsedbackwardkernel}), we obtain the so-called incremental weights
\begin{equation}\label{eq:incrementalweightSMC}
    G_k(x_{k-1})=\frac{\gamma_k(x_k)L^{\textup{col}}_{k-1}(x_k,x_{k-1} )}{\gamma_{k-1}(x_{k-1})M^{\textup{col}}_k(x_{k-1},x_k)}=\frac{\unnormDensity_{k}(\baseFlow_{\timeIndex}(\x_{\timeIndex-1}))}{\unnormDensity_{\timeIndex-1}(\x_{\timeIndex-1})}|\nabla \baseFlow_{\timeIndex}(\x_{\timeIndex-1}) | 
\end{equation}
as given in (\ref{eq:GDef}).
In \cite{wunoe2020stochastic}, the weight $w_K(x_{0:K-1})$ is instead obtained by first computing recursively the product appearing on the RHS of (\ref{eq:decompoweightSNF}), thus they consider formally\footnote{Note that while $G_k(x_{k-1})$ corresponds to the Radon-Nikodym derivative between the measures $\gamma_k(\mathrm{d}x_k)L^{\textup{col}}_{k-1}(x_k,\mathrm{d}x_{k-1})$ and  $\gamma_{k-1}(\mathrm{d}x_{k-1})M^{\textup{col}}_{k}(x_{k-1},\mathrm{d}x_{k})$, the ratio  $L^{\textup{col}}_{k-1}(x_{k}, \mathrm{d}x_{k-1})/M^{\textup{col}}_k(x_{k-1}, \mathrm{d}x_k)$ is not defined and taking its logarithm can only be interpreted as a shorthand notation for $\log G_k(x_{k-1})-\log (\gamma_k(x_k)/\gamma_{k-1}(x_{k-1}))$.} in equation (12) of their paper a log-weight update at iteration $k$ of the form
\begin{align}
    \log \Big(\frac{L^{\textup{col}}_{k-1}(x_{k}, x_{k-1})}{M^{\textup{col}}_k(x_{k-1}, x_k)}\Big)&=\log (\gamma_k(T_k(x_{k-1})|\nabla T_l(x_{l-1})|)-\log \gamma_k(x_k)\nonumber\\
    &=\log \unnormDensity_{\timeIndex}(\x^{\text{SNF}}_{\timeIndex}) - \log \unnormDensity_{\timeIndex}(\hat{\x}^{\text{SNF}}_{\timeIndex})
\end{align}
where $\hat{\x}^{\text{SNF}}_{\timeIndex}$ is the value of the sample after the MCMC at temperature $\timeIndex$ has been applied and $\x^{\text{SNF}}_{\timeIndex}$ is the value of the sample before it as described in (\ref{eq:updateweightSNF}).

\section{Gradient expression for ELBO based approaches}\label{section:elboGradient}
Assume we consider the ELBO objective as in \cite{wunoe2020stochastic}, that is we are interested in maximizing w.r.t. to the NFs parameters and the stochastic kernel parameters
\begin{align}
\mathcal{L}=\mathbb{E}_{X_{0:K}\sim \bar{\eta}_K}\Big[\log w_K(X_{0:K-1})\Big],
\end{align}
where $\bar{\eta}(x_{0:K})=\bar{\eta}_K(x_{0:K})$ is defined in \eqref{eq:extendedproposalcollapsednoresampling} and, as shown in (\ref{eq:productincrementalweight}), we have 
\begin{equation}
 w_K(x_{0:K-1})=\frac{Z_K \bar{\pi}_K(x_{0:K})}{\bar{\eta}_K(x_{0:K})}=\prod_{l=1}^K G_l(x_{l-1})\\
\end{equation}
for $\bar{\pi}_K(x_{0:K})$ defined in \eqref{eq:extendedtargetcollapsednoresampling}. It is thus trivial to check that $w_K(X_{0:K-1})$ is an unbiased estimate of $Z_K$ when  $X_{0:K} \sim \bar{\eta}_K$. 

\citet{Thin:2021} provide a gradient estimate of the ELBO for annealed importance sampling, which corresponds to the case where $T_l(x)=x$ for all $l$, when the MCMC kernels $\mathcal{K}_l$ are of the Metropolis--Hastings type and we use a reparameterized proposal 
\begin{equation}\label{eq:MHkernel}
    \mathcal{K}_l(x,x')=\int Q_l((x,u),x') g(u)\mathrm{du}
\end{equation}
for 
\begin{equation}\label{eq:MHmap}
   Q_l((x,u),x') =\alpha_l(x,u) \delta_{S_l(x,u)}(x')+\{1-\alpha_l(x,u)\} \delta_{x}(x'),
\end{equation}
We show here how one can directly extend these results to our settings. 
We will use the convention $\alpha^1_l(x,u)=\alpha_l(x,u)$, $S^1_l(x,u)=S_l(x,u)$, $\alpha^0_l(x,u)=1-\alpha_l(x,u)$ and $S^0_l(x,u)=x$ so that we can rewrite 
\begin{equation}
   Q_l((x,u),x') =\sum_{a_l=0}^1 \alpha^{a_l}_l(x,u) \delta_{S^{a_l}_l(x,u)}(x').
\end{equation}
By combining  \eqref{eq:collapsedkernel}, \eqref{eq:extendedproposalcollapsednoresampling}, \eqref{eq:MHkernel} and \eqref{eq:MHmap}, we can rewrite the distribution $\bar{\eta}_k(x_{0:K})$ as 
\begin{align}
    \bar{\eta}_K(x_{0:K})    &=\pi_0(x_0) \prod_{l=1}^K \mathcal{K}_l(T_l(x_{l-1}),x_l) \nonumber\\
    &=\pi_0(x_0) \prod_{l=1}^K \int Q_l((T_l(x_{l-1}),u_l),x_l)g(u_l)\mathrm{d}u_l \nonumber\\
    &=\pi_0(x_0) \int \cdots \int \sum_{a_{1:K}}  \prod_{l=1}^K \Big(g(u_l) \alpha^{a_l}_l(T_l(\Phi_{l-1}(x_0,u_{1:l-1},a_{1:l-1})),u_l)  \delta_{S^{a_l}_l(T_l(\Phi_{l-1}(x_0,u_{1:l-1},a_{1:l-1})),u_l)}(x_l) \Big)\mathrm{d}u_{1:K}\nonumber\\
    &=\pi_0(x_0) \int \cdots \int \sum_{a_{1:K}} g(u_{1:K}) \beta(a_{1:K}|x_0,u_{1:K}) \prod_{l=1}^K \delta_{S^{a_l}_l(T_l(\Phi_{l-1}(x_0,u_{1:l-1},a_{1:l-1})),u_l)}(x_l) \mathrm{d}u_{1:K}
\end{align}
where, for a given realization of $u_{1:L}$, we draw sequentially the Bernoulli r.v. $A_l$ with 
\begin{equation}
    \mathbb{P}(A_l=a_l|x_0,u_{1:l-1},a_{1:l-1})=\alpha^{a_l}_l(T_l(x_{l-1}),u_l)
\end{equation}
and, given $x_0,u_{1:l},a_{1:l}$, the state $x_l$ is given by
\begin{equation}\label{eq:xdeterministic}
x_l:=\Phi_l(x_0,u_{1:l},a_{1:l})=S^{a_l}_l(T_l(x_{l-1}),u_l)=S^{a_l}_l(T_l(\Phi_{l-1}(x_0,u_{1:l-1},a_{1:l-1})),u_l)
\end{equation}
with $\Phi_0(x_0,u_{1:0},a_{1:0}):=x_0$. Finally we used the notation $g(u_{1:K})=\prod_{l=1}^K g(u_l)$ and the joint distribution of $A_{1:K}$ is denoted 
\begin{equation}
    \beta(a_{1:K}|x_0,u_{1:K})=\prod_{l=1}^K \alpha^{a_l}_l(T_l(\Phi_{l-1}(x_0,u_{1:l-1},a_{1:l-1})),u_l).
\end{equation}

So we can rewrite the ELBO as 
\begin{align}
    \mathcal{L}&=\mathbb{E}_{\bar{\eta}_K(x_{0:K})}\Big[\log w_K(x_{0:K-1})\Big]\nonumber\\
    &=\mathbb{E}_{\pi_0(x_0)g(u_{1:K})\beta(a_{1:K}|x_0,u_{1:K})}\Big[\log w_K(x_{0:K-1})\Big],
\end{align}
where we recall that the states $x_{0:K}$ are deterministic functions of $x_0,u_{1:K},a_{1:K}$ given by \eqref{eq:xdeterministic}. Now when taking the gradient of the ELBO w.r.t. parameters of the NFs and/or stochastic kernels then we have 
\begin{equation}\label{eq:gradient_elbo_appendix}
    \nabla \mathcal{L}=\mathbb{E}_{\pi_0(x_0)g(u_{1:K})\beta(a_{1:K}|x_0,u_{1:K})}\Big[\nabla \log w_K(x_{0:K-1}) + \nabla \log \beta(a_{1:K}|x_0,u_{1:K}) \cdot \log w_K(x_{0:K-1}) \Big].
\end{equation}
Now set  $u_0 {=} x_0$, $u{=}u_{0:K}$, $u{=} a_{1:K}$. Hence, the random variables $(A,U)$ admit a joint distribution $p$  of the form 
$p(a,u) {:=}\pi_0(u_0)g(u_{1:K})\beta(a_{1:L}|x_0,u_{1:K})$. Moreover, define $\Phi$ the re-parametrization that maps $(A,U)$ to the samples $X_{0:K}$, i.e. $[\Phi(A,U)]_l=\Phi_l(U_{0:l},A_{1:l})=X_l$. Hence, we can express the gradient in \cref{eq:gradient_elbo_appendix} as follows:
\begin{align}
		\expect_{p}\left[ \nabla \log (\particleWeightUnnorm_{\numTransitions}\circ \Phi) + \log(\particleWeightUnnorm_{\numTransitions}\circ \Phi) ~ \nabla \log p  \right].	
\end{align}

The first term on the r.h.s. corresponds to the reparametrization trick while the second term is a score term, which in general can have high variance.

The gradient estimator of \citep{wunoe2020stochastic} effectively uses an unbiased estimator of the first reparameterization term but neglects the second term altogether.

In the particular case where the flows transport exactly between each temperature, i.e. $T_l^{\#}\pi_{l-1}{=}\pi_l$, the importance weight $w_K(x_{0:K-1})$ has the constant value $Z_K$.  Hence, the second term in the expectation is a constant multiplied by a score function and consequently it vanishes:
\begin{align}
		\expect_{p}\left[ \log (\particleWeightUnnorm_{\numTransitions}\circ g) \nabla \log p  \right] 
		\propto 
		\expect_{p}\left[  \nabla \log p  \right]	= \int \nabla p = 0.
\end{align}
Moreover, since the ELBO is maximized when using optimal flows, the gradient $\nabla \mathcal{L}$ must vanish, which directly implies that the first term must also vanish. Therefore, despite the missing term in the gradient, the SNF learning rule has the correct fixed point in expectation.

\section{Particle Interpretation of CRAFT}\label{sec:appparticleintepretationCRAFT}
CRAFT is attempting to minimize w.r.t. $\theta$ the KL divergence $\mathcal{KL}(T^{\pushForward}_k(\theta)\pi_k||\pi_{k+1})=\mathcal{KL}(\pi_k||(T^{-1}_k(\theta))^{\pushForward}\pi_{k+1})$ We have shown in Section \ref{section:particleKL} that we could re-interpret the criterion minimized by CRAFT at time $k$ as the KL divergence between the expectation of the random measure approximating $\pi_k$ and the pullback of $\pi_{k+1}$ by $T_k(\theta)$. This was established in the case where the random measure is obtained using importance sampling. We show here that it is also valid for the case considered by CRAFT where it arises from an SMC-NF algorithm.

For sake of simplicity, we assume resampling is performed at any time step. Then in this case, the particles generated by the SMC-NF algorithm are distributed according to 
\begin{align}\label{eq:distributionparticlesystem}
    \bar{q}_k(\mathrm{d}x^{1:N}_{0:k},\mathrm{d}y^{1:N}_{1:k})=\prod_{i=1}^N \pi_0(\mathrm{d}x^i_0)  \prod_{l=1}^{k} \Big[\prod_{i=1}^N \delta_{T_l(x^i_{l-1})}(\mathrm{d}y^i_l) \prod_{i=1}^N W^{a^i_{l-1}}_{l}  \mathcal{K}_{l}(y^{a^i_{l-1}}_{l},\mathrm{d}x^i_l)\Big],
\end{align}
where $a_{l-1}^i\in\{1,...,N\}$ is the ancestral index of particle $x^l_i$. Using random particles arising from the SMC-NF algorithm, the distribution $\pi_k$ is then approximated by the random empirical measure 
\begin{equation}
\pi^N_{k}(\mathrm{d}\x)=\frac{1}{N}\sum_{i=1}^N ~\delta_{X^i_k}(\mathrm{d}x)
\end{equation}
and its expectation w.r.t. (\ref{eq:distributionparticlesystem}) is denoted
\begin{equation}
    \hat{\pi}^N_k(\mathrm{d}\x)=\mathbb{E}[\pi^N_{k}(\mathrm{d}\x)].
\end{equation}
In this case, we have 
\begin{align}
    \mathcal{KL}(\hat{\pi}^N_k || T^{-1}_k(\theta)_{\pushForward}\pi_{k+1})&=\mathbb{E}_{X \sim \hat{\pi}^N_k}\bigg[\log \bigg( \frac{\hat{\pi}^N_k(X)}{(T^{-1}_k(\theta))^{\pushForward}\pi_{k+1}(X)}\bigg) \bigg]\nonumber\\
    &=\mathbb{E}_{(X^{1:N}_{0:k},Y^{1:N}_{1:k}) \sim \bar{q}_k, X \sim \pi^N_k}\bigg[\log \bigg( \frac{\hat{\pi}^N_k(X)}{(T^{-1}_k(\theta))^{\pushForward}\pi_{k+1}(X)}\bigg) \bigg]\nonumber\\
    &=\mathbb{E}_{(X^{1:N}_{0:k},Y^{1:N}_{1:k}) \sim \bar{q}_k}\bigg[ \mathbb{E}_{X \sim \pi^N_k}\bigg[\log \bigg( \frac{\hat{\pi}_k(X)}{(T^{-1}_k(\theta))^{\pushForward}\pi_{k+1}(X)}\bigg) \bigg]\bigg].
\end{align}
So an unbiased gradient estimate w.r.t. $\theta$ of $\mathcal{KL}(\hat{\pi}_k || T^{-1}_k(\theta)_{\pushForward}\pi_{k+1})$ is given by 
\begin{equation}
-\mathbb{E}_{X \sim \pi^N_k}\bigg[\nabla_\theta \log \Big({T^{-1}_k(\theta)^{\pushForward}\pi_{k+1}}\Big) \bigg],
\end{equation}
which is the gradient used by CRAFT to update the parameter of $T_k(\theta)$.

\section{Particle MCMC}\label{sec:particleMCMCdescription}
We describe here the particle independent Metropolis--Hastings algorithm (PIMH) introduced in \citep{Andrieu2011}. This is an MCMC algorithm using as an independent proposal an importance sampling or SMC-type  algorithm such as an SMC combined with NFs learned by CRAFT (see Algorithm \ref{algo:smcFlow}). It generates a Markov chain $((\targetDist^\numParticles_\numTransitions(j),\normConst_\numTransitions^\numParticles(j))_{j\geq 1}$ where $\targetDist^\numParticles_\numTransitions(j)(\mathrm{d}x)=\sum_{i=1}^\numParticles \particleWeightNorm^i_{\numTransitions}(j) \delta_{\particleX^i_\numTransitions(j)}(\mathrm{d}x)$ such that for any test function $f$
\begin{equation}\label{eq:MCMCestimator}
    \frac{1}{J}\sum_{j=1}^J \mathbb{E}_{X\sim \targetDist^\numParticles_\numTransitions(j)}[f(X)] \rightarrow \mathbb{E}_{X\sim \targetDist_\numTransitions}[f(X)],\quad\text{where}\quad \mathbb{E}_{X\sim \targetDist^\numParticles_\numTransitions(j)}[f(X)]:=\sum_{i=1}^\numParticles \particleWeightNorm^i_\numTransitions(j)f(\particleX^i_\numTransitions(j)),
\end{equation}
as the number $J$ of MCMC iterations goes to infinity. This is described in detail in Algorithm \ref{algo:PIMH}.  
\begin{algorithm}
\caption{Particle Independent Metropolis--Hastings step  $P((\targetDist^\numParticles_\numTransitions(j),\normConst_\numTransitions^\numParticles(j)),(\cdot,\cdot))$\label{alg:pimh}}\label{algo:PIMH}
\begin{algorithmic}[1]
\STATE \textbf{Input:} Current approximations $(\targetDist^\numParticles_\numTransitions(j),\normConst_\numTransitions^\numParticles(j))$ to $(\targetDist_\numTransitions,\normConst_\numTransitions)$.

\STATE \textbf{Output:} New approximations $(\targetDist^\numParticles_\numTransitions(j+1),\normConst_\numTransitions^\numParticles(j+1))$ to $(\targetDist_\numTransitions,\normConst_\numTransitions)$.

\STATE Run an SMC algorithm to obtain $(\targetDist^{\star,\numParticles}_\numTransitions,\normConst^{\star,\numParticles}_\numTransitions)$ approximating $(\targetDist_\numTransitions,\normConst_\numTransitions)$ %
(such as  Algorithm \ref{algo:smcFlow}).
\STATE  Set $(\targetDist^\numParticles_\numTransitions(j+1),\normConst_\numTransitions^\numParticles(j+1))=(\targetDist^{\star,\numParticles}_\numTransitions,\normConst^{\star,\numParticles}_\numTransitions)$ with probability $\alpha(\normConst_\numTransitions^\numParticles(j),\normConst^{\star,\numParticles}_\numTransitions):=\min \Big\{1, \frac{\normConst^{\star,\numParticles}_\numTransitions}{\normConst_\numTransitions^\numParticles(j)}\Big\}.$
\STATE  Otherwise set $(\targetDist^\numParticles_\numTransitions(j+1),\normConst_\numTransitions^\numParticles(j+1))=(\targetDist^\numParticles_\numTransitions(j),\normConst_\numTransitions^\numParticles(j).$
\end{algorithmic}
\end{algorithm}

For fixed computational efforts, one could either run many MCMC iterations with $N$ small or few MCMC iterations with $N$ large. Under regularity conditions, it was shown in \cite{pitt2012some} that $N$ should be selected such that the variance of the estimate of the log-normalizing constant is close to $1$ in order to minimize the asymptotic variance of the estimator (\ref{eq:MCMCestimator}). 

\section{Learning model parameters using CRAFT}\label{sec:learningmodelparameters usingCRAFT}
We do not consider in the paper examples where $\targetDist_\numTransitions$ depends on some parameters $\phi$ that one wants to learn. We explain here how this could be addressed. 

Consider the following target 
\begin{equation}
    \targetDist^{\phi}_\numTransitions(x)=\frac{\unnormDensity^{\phi}_\numTransitions(x)}{\normConst^{\phi}_\numTransitions}.
\end{equation}
To estimate $\phi$, we propose to learn $\phi$ by maximizing the log-normalizing constant/evidence 
\begin{equation}
    \arg \min_{\phi} \log \normConst^{\phi}_\numTransitions.
\end{equation}
The gradient of this objective w.r.t. $\phi$ can be approximated as follows 
\begin{equation}
    \nabla \log \normConst^{\phi}_\numTransitions =\mathbb{E}_{X \sim \targetDist^{\phi}_\numTransitions}[\nabla \log \unnormDensity^{\phi}_\numTransitions(X)] \approx \mathbb{E}_{X \sim \targetDist^{N,\phi}_\numTransitions}[\nabla \log \unnormDensity^{\phi}_\numTransitions(X)],
\end{equation}
where $\targetDist^{N,\phi}_\numTransitions$ is an particle approximation of $\targetDist^{\phi}_\numTransitions$. This approximation can be obtained using a learned SMC/normalizing flow sampler obtained using CRAFT.

\section{Further comparison of CRAFT and AFT}\label{section:CRAFTvsAFTextra}

In this section we investigate the effect of reducing the number of MCMC iterations in CRAFT and AFT. As a benchmark we use the log Gaussian Cox process- with a large lattice discretization of $40 \times 40 = 1600$. To make the example harder we reduce the number of HMC iterations per temperature from $10$ to $2$. The rest of the experimental setup is kept the same. The goal here is to learn a fast flow augmented sampler at test time with reasonable training time. Since the diagonal affine flow used constitutes a negligible overhead the change corresponds to a $5$ times reduction in test compute time relative to the previous work. 

\begin{figure}[ht]
\vskip 0.2in
\begin{center}
\centerline{\includegraphics[width=0.9\linewidth]{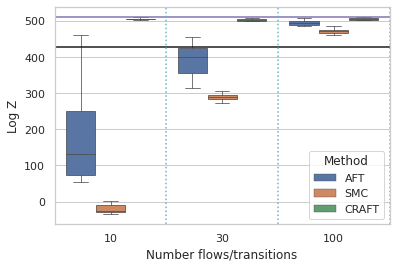}}
\caption{Comparison of CRAFT, AFT, SMC and VI on the pines task from \citep{Arbel2021} with a reduced number of HMC iterations.}
\label{fig:pinesAFTvsCRAFT}
\end{center}
\vskip -0.2in
\end{figure}

Figure \ref{fig:pinesAFTvsCRAFT} shows the results. We see that for $10$ and $30$ transitions that AFT struggles to learn a good flow. This is because the expectation of the objective required in the greedy training is not well estimated with fewer MCMC iterations. CRAFT has the same difficulty at the beginning of optimization but due to the repeated nature of the optimization it is able to recover. 
Having seen the solution found by CRAFT one might reasonably ask if we could use different numbers of HMC iterations at train and test time for AFT- though this is not suggested in the original work of \citep{Arbel2021}.  After all, we know that with 10 iterations AFT performs well on the Cox process task, which suggests we could use that learnt flow with fewer HMC samples at test time. This is indeed the case. Training with extra MCMC iterations restores the behaviour of AFT to be similar to that of CRAFT in this case. However this is at the expense of introducing even more hyperparameters to AFT that require tuning. Relative to CRAFT, a user of AFT has to correctly tune the behaviour of three sets of particles (train/validation/test) and manage multiple separate optimization loops (one for each temperature). This extra modification of allowing different number of MCMC iterations is then a further complexity. 

Note additionally that AFT is subject to the degradation in performance with batch size relative to CRAFT described in Section \ref{section:CRAFTvsAFT}- we regard this case as the most direct demonstration of how CRAFT solves the sample replenishment problem in AFT.

\section{Distributed and parallel computation}

 In this section we discuss the particular challenges we envisage scaling the algorithm to large scale compute infrastructure and potential remedies.

We have described the algorithm as a sequential algorithm but there is considerable scope for running it asynchronously in parallel. Each temperature can be considered to correspond to a node in a chain graph. Along the edges of the graph messages are passed. In the most direct implementation this corresponds to the particles and the weights. There is no need to run for a full pass along the graph before starting the next one. In some compute clusters communication locality is also important so it is helpful that the message passing is sparse and local.

An alternative to passing the particles and weights is to cycle the temperature at each node and pass gradient updates instead. The temperatures at each node can be offset by one so that the gradients are always received at the right time. Such a prescription is inspired by the distributed Algorithm 5 of \citep{Syed2019}. For CRAFT, this method will reduce communication overhead between nodes in the case   
where the gradient updates require less memory than passing particles and weights. Although the gradient updates for neural networks can in general be large, incorporating symmetries as in Sections \ref{section:symmetry} and \ref{section:phiFour} or more generally adding domain knowledge can reduce network sizes substantially. 

Another consideration is the case where only a few particles will fit onto a worker, so that it is desirable to parallelize a single iteration across multiple workers. The resampling step means that the algorithm is not batch parallel. Therefore resampling could introduce a significant communication overhead between the parallel workers. One possible remedy is only to resample within a worker.  %

\section{Incorporating symmetries}\label{section:symmetry}

 In applications to high dimensional target densities the target density often exhibits symmetries. Incorporating such symmetries in normalizing flows can give significant improvements in the performance and efficiency of normalizing flows. 

Consider a case where we have a group $\genericGroup$ with elements $\genericGroupElement$ and corresponding group representation members $\genericGroupRepresentationElement_{\genericGroupElement}$. In this case we have a symmetric target density $\flowTargetDensity$ which is symmetric under the target density.

\begin{equation}
\flowTargetDensity(\genericGroupRepresentationElement_{\genericGroupElement}\x) = \flowTargetDensity(\x) \hspace{5 pt} \forall \genericGroupElement \in \genericGroup
\end{equation}

We also take a base density $\flowBaseDensity$ that is also symmetric under the group. We choose a normalizing flow that is equivariant under the action of the group so that:

\begin{equation}
\baseFlow(\genericGroupRepresentationElement_{\genericGroupElement}\x) = \genericGroupRepresentationElement_{\genericGroupElement} \baseFlow (\x)\hspace{5 pt} \forall \genericGroupElement \in \genericGroup
\end{equation}

When we push forward the base density we end up with a distribution $\baseFlow_{\pushForward}\flowBaseDensity$ that is invariant under the action of the group.

\begin{equation}
\baseFlow_{\pushForward}\flowBaseDensity(\genericGroupRepresentationElement_{\genericGroupElement}\x) = \baseFlow_{\pushForward}\flowBaseDensity(\x) \hspace{5 pt} \forall \genericGroupElement \in \genericGroup
\end{equation}

In standard normalizing flow modelling the symmetric pushforward distribution is then tuned to match the symmetric target distribution using reverse $\KL$-divergence $\KL[\baseFlow_{\pushForward}\flowBaseDensity || \flowTargetDensity]$.

It is important to extend such a treatment of symmetries to the case of CRAFT. To this end we analyze the symmetry properties of a sequence of densities with a geometric annealing schedule. We consider an initial normalized  density $\initialDensity$ and unnormalized final density $\finalUDensity$. The unnormalized density of the geometric annealing schedule takes the form:

\begin{equation}
\log \annealedUDensity(\x) = \log (1-\beta) \log \initialDensity(\x) + \beta \log \finalUDensity(\x)    
\end{equation}

\noindent where the schedule time $\beta \in [0, 1]$. Clearly if both the initial density and final density are symmetric under the action of the group then so is $\annealedUDensity(\x)$ and its normalized version $\annealedDensity$.

In CRAFT we use a sum of $\KL$ divergences between the pushforward of an initial annealed density and the next target density $\KL[\baseFlow \pushForward\annealedDensity || \pi_{\beta'} ]$. In terms of symmetry there is a direct analogy to the case of standard variational inference with normalizing flows with which we started the discussion. In particular $\annealedDensity$ now takes the role of the symmetric base density and $\pi_{\beta'}$ takes the role of the target. The symmetry requirement on the normalizing flow $\baseFlow$ is thus the same. It needs to be equivariant under the action of the group so that the push forward $\baseFlow \pushForward\annealedDensity$ is symmetric in a manner that matches its target.

\section{Additional experiment details}

All experiments used a geometric (log-linear) annealing schedule. The initial distribution was always a standard multivariate normal. All experiments used HMC as the Markov kernel, which was tuned to get a reasonable acceptance rate based on preliminary runs of SMC. Normalizing flows were always initialized to the identity flow. Wherever a stochastic gradient optimizer was required we used the Adam optimizer \citep{Kingma:2014}.

In terms of software dependencies for our code we used Python, JAX \cite{Bradbury:2018}, Optax \cite{optax2020github}, Haiku \cite{Hennigan:2020},  and the TensorFlow probability JAX `substrate' \citep{Dillon:2017}.

All experiments were carried out using a single Nvidia v100 GPU. We used sufficient CPUs and CPU RAM such that these were not the bottleneck. 

\subsection{Comparing CRAFT and AFT batch size performance}

Here we give more details of the experiment described in Section \ref{section:CRAFTvsAFT}. We used the original software for AFT which is available at \url{https://github.com/deepmind/annealed_flow_transport}. We used practical AFT (Algorithm \ref{algo:AFTpractical}), which is the version using early stopping on a set of validation particles and a `hold out' test set. 

We used the same MCMC method for both approaches, which was pre-tuned full Metropolis adjusted Hamiltonian Monte Carlo (HMC) with 10 leapfrog steps per iteration. For CRAFT we used 10 HMC steps per temperature and for AFT we used 1000 HMC steps per temperature meaning that AFT had 100 times more HMC steps per temperature. The HMC step sizes were linearly interpolated with $0$ corresponding to the initial distribution and $1$ corresponding to the final distribution, the interpolation points were $[0, 0.25, 0.5. 1]$ and the corresponding step sizes were $[0.3, 0.3, 0.2, 0.2]$.

At test time AFT and CRAFT used the same number of particles for each experiment. To make it fair at train time, the total CRAFT particle budget was divided in to two halves for AFT, one half was used for the training particles and the other half was used for the validation particles. For each experiment the total number of train particles for either method was the same as the number used at test time and this is the number shown in Figure \ref{fig:AFTvsCRAFT}.

In terms of optimization both methods were well converged. AFT used an optimization step size of 1e-2, and $500$ optimization iterations per temperature. CRAFT used $200$ total optimization iterations, and a step size of 5e-2 which was reduced down to 1e-2 after $100$ iterations.

\subsection{Comparing CRAFT and Stochastic Normalizing Flows}

To the best of knowledge, the HMC based Stochastic Normalizing Flow described in \citep{wunoe2020stochastic} was not implemented or experimented with in that work. We implemented it and found in preliminary experiments that it outperformed the random walk Metropolis Hastings used in the original work. It was also the most commensurate with our own HMC usage. Therefore we used the HMC SNF for the comparisons in this work.

In the spirit of the original SNF paper, we performed preliminary experiments with a SNF ELBO that incorporated resampling in the forward computation. For the reverse computation we neglected the resampling contribution just as is done for the score term arising from the discrete Metropolis--Hastings correction. In other words we proceeded as if the forward computation was compatible with the reparameterization gradient although this does not in fact correspond to an unbiased estimate of the gradient. These preliminary experiments indicated that this variant of SNF became unstable and challenging to train. The fact that CRAFT can cope with resampling is a benefit of the approach, and is expected from the form of the CRAFT training objective. 

We observed that variational learning of MCMC step sizes is unstable in SNF, so we followed \citep{wunoe2020stochastic} who use a fixed step size during variational training for the majority of their experiments. Instead of a learnt step size, we tuned the step sizes separately in advance so that the corresponding MCMC would have a reasonable acceptance rate just as we do for CRAFT. Consequently we used the same Markov kernels for both CRAFT and SNFs.

For the SNF software we implemented them in JAX using similar modules to that of the CRAFT implementation. We verified this JAX implementation against the original publicly available SNF implementation at \url{https://github.com/noegroup/stochastic_normalizing_flows}. A benefit of using the same basic modules as the CRAFT implementation is that the timed comparison is much less subject to unwanted differences arising from the different software libraries used. We confirmed using a Python execution profiler that both the CRAFT and SNF code was spending time on the expected core algorithmic computations and not on unexpected code paths. We observed some variability in the CRAFT and SNF step times.   

The JAX SNF implementation exploits the connection between SNFs and Annealed Importance Sampling with normalizing flows. This equivalence can be easily seen from the form of the overall unnormalized weights in each case (Appendix \ref{sec:appextendedproposaltargetSNF}). The gradient dynamics are unchanged by using this representation of the forward computation. 

Relative to CRAFT we found that SNFs used significantly more memory in our experiments. This is due to the backward pass of the SNFs where gradients are passed through the whole forward computation whereas the gradients are local in CRAFT. To the benefit of SNFs, we reduced the batch size of both CRAFT and SNFs in the relevant experiments so that the SNF training would still fit on GPU and that the batch sizes would be comparable for the two algorithms. 

The normalizing flow used for the VAE experiment was an Affine Inverse Autoregressive Flow. Similiar to \citep{Arbel2021} we use a masked network \citep{Germain:15} to achieve the autoregressive dependency structure. The masked network had two hidden layers and unmasked would have 150 hidden units. The final weights and biases were initialized to give an identity flow.  We used a leaky ReLU nonlinearity. Since the flows have high capacity in this case we only needed to use 3 temperatures in both cases. Since the cost of these inverse autoregressive flows is quadratic in the dimension they are prohibitively expensive for systems of higher dimensionality. We used 2 HMC steps per transition with 10 leapfrog steps in both cases. Both CRAFT and SNF had a batch size of 100. 

For the LGCP SNF/CRAFT comparison we used 10 transitions in both cases. We used 1 HMC step per transition with 10 leapfrog steps in both cases. Both CRAFT and SNF had a batch size of 2000. 

\subsection{CRAFT based Particle MCMC for lattice $\phi^4$ theory}

Those readers who are less familiar with the background physics can gain intuition from the function form of equation \eqref{eq:latticePhiFour}. The terms over neighbouring lattice sites promote spatial correlation. The other terms all involve only a single lattice site. 

\begin{figure*}[ht!]
\begin{center}
\centerline{\includegraphics[width=0.95\linewidth]{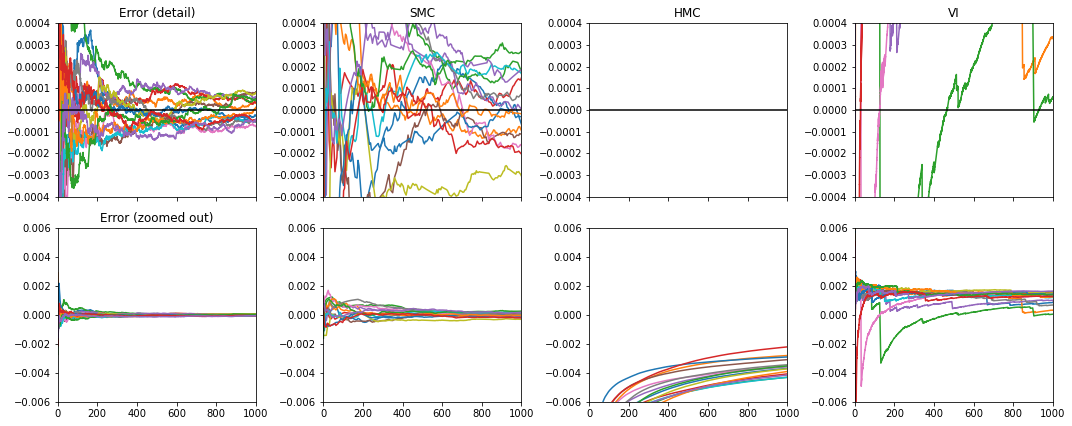}}
\caption{Timed comparison of MCMC methods for the $\phi^4$ example, based on fifteen repeats. CRAFT, SMC and VI serve as proposal mechanisms for particle MCMC. HMC is applied directly to the target. Note HMC never reaches the detailed level of error in the top row.}
\label{fig:phiFourComparisonIsing}
\end{center}
\end{figure*}

\begin{figure*}[ht!]
\vskip 0.2in
\begin{center}
\centerline{\includegraphics[width=0.95\linewidth]{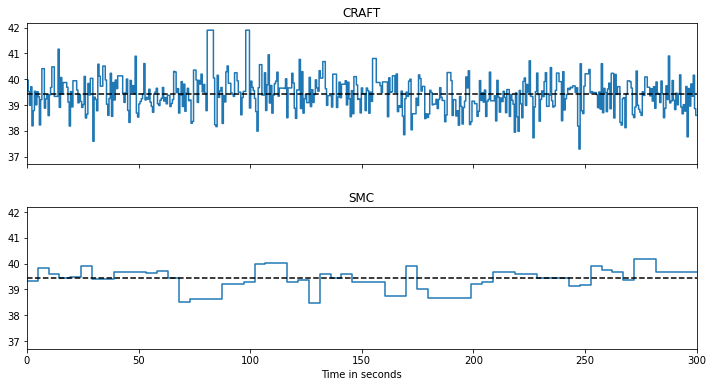}}
\caption{Five minutes of the raw chain used to compute the expectation averages for the SMC and CRAFT two point susceptibility. The estimated gold standard value is shown as the black dashed line. SMC and CRAFT are displayed because they are the two best algorithms. VI and HMC have much larger errors.}
\label{fig:twoPointRawChain}
\end{center}
\end{figure*}

\begin{figure*}[ht!]
\vskip 0.2in
\begin{center}
\centerline{\includegraphics[width=0.95\linewidth]{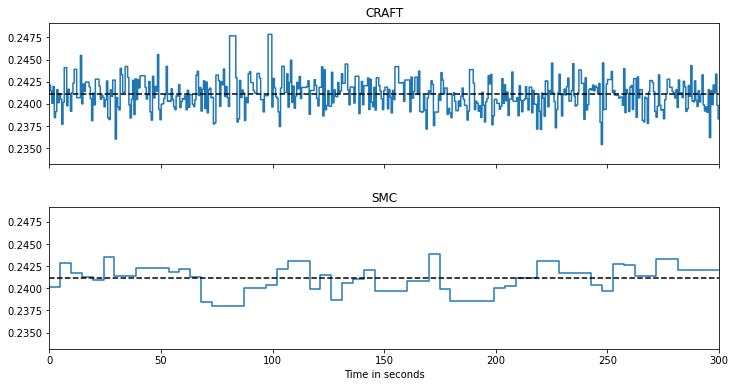}}
\caption{Five minutes of the raw chain used to compute the expectation averages for the SMC and CRAFT Ising energy density. The estimated gold standard value is shown as the black dashed line. SMC and CRAFT are displayed because they are the two best algorithms. VI and HMC have much larger errors.}
\label{fig:isingRawChain}
\end{center}
\end{figure*}

The expectations we use to evaluate the algorithms are physically motivated observables described by \citep{Albergo2019}. In the main text we use the two point susceptibility. In Figure \ref{fig:phiFourComparisonIsing} we show that similar results show for the Ising energy density. The chains that were averaged to produce the expectations in \ref{fig:phiFourComparison} and \ref{fig:phiFourComparisonIsing} are shown in \ref{fig:twoPointRawChain} and \ref{fig:isingRawChain} respectively. The errors shown in the average value plots are computed by subtracting off the gold standard value.

We note that \citet{DeHaan2021} incorporate additional symmetries of the target on top of the lattice translation symmetry we incorporate. This is at the expense of introducing numerical integration into the variational inference with normalizing flows. The convolutional affine coupling layers we use do not require such numerical integration, and are known to significantly outperform fully connected affine coupling layers in this context \citep{Albergo2021}. 

The parameters we consider for the theory are based on those from \citep{Albergo2019}. In particular we take the hardest parameter set E5 and make it more difficult by reducing $m^2$ from $-4$ to $-4.75$. As noted by \citep{Albergo2019}, the reason this makes the problem more challenging is it increases the gap between the positive and negative modes. The resulting parameters we use are therefore the following: Lattice size $14\times 14$; $\coupling=5.1$ and $m^2=-4.75$. 

In these $\phi^4$ experiments, upon preliminary investigation of computation time using a Python execution profiler, we found that time was sometimes being spent in parts of the code that might not be expected from algorithmic considerations alone. In particular for fast samplers like VI, we found that storing results and computing expectations were an (unoptimized) bottleneck. To make timed comparisons fairer, we therefore performed separate timing estimation on code that ran only the core sampling code and not on computing expectations and storing results. These adjusted timing estimates are what is reported in our results.

The gold standard values of the error for each reported expectation are computed using SMC. We verified in cases of simple known expectations that SMC and CRAFT gave the correct answers whereas HMC and VI gave large errors, just as for the physically motivated observables reported.

The normalizing flow used a convolutional affine coupling layer \citep{Dinh2017} with periodic boundary conditions. We used alternating checkerboard masking patterns. The convolution kernels were of size $3\times 3$. The conditioner of each flow had one hidden convolutional layer with 10 channels. The final weights and the biases of the convolution were initialized to give an identity flow. We used a ReLU nonlinearity.    

SMC used 90 transitions.  Since the masking pattern required two coupling layers to change each site we used two coupling layers per transition in CRAFT. CRAFT used 10 transitions. VI, which has no annealing, used 20 coupling layers. We confirmed there was no benefit to increasing the number of VI coupling layers. We used 10 HMC steps per annealing transition with 10 leapfrog steps for SMC and CRAFT. Direct HMC used 10 leapfrog iterations per step.

\clearpage

\section{CRAFT to AFT algorithm comparison.}

In this section we describe the two different variants of AFT from \citep{Arbel2021} for clarity and completeness. We also describe the line difference between CRAFT and simple AFT. Note that the full source code for each method is also available at~\url{https://github.com/deepmind/annealed_flow_transport}. Based on equation (\ref{eq:lHat}) we define:

\begin{equation}\label{eq:lHatAlgo}
\mathcal{L}^N_k(T) :=  \KL[\baseFlow^{\#}_{\timeIndex} \targetDist_{\timeIndex-1} || \targetDist_{\timeIndex}] 
 - \log \left(\frac{\normConst_\timeIndex}{\normConst_{\timeIndex-1}}\right) \approx \sum_{\particleIndex} \particleWeightNorm_{\timeIndex-1}^{\particleIndex} \deltaDensity_{\timeIndex}(\particleX_{\timeIndex-1}^{\particleIndex})
\end{equation}

which from equation \eqref{eq:gradientApproximation} has approximate gradient given by

\begin{equation}\label{eq:gradientApproximationAlgo}
\sum_{\particleIndex} \particleWeightNorm_{\timeIndex-1}^{\particleIndex} \frac{\partial \deltaDensity_{\timeIndex}(\particleX_{\timeIndex-1}^{\particleIndex})}{\partial \genericFlowParams_{\timeIndex}}.
\end{equation}

\begin{algorithm}
\caption{Simple AFT: from Arbel, Matthews and Doucet 2021. This corresponds to Algorithm 1 of the prior paper.}\label{algo:AFTsimple} 	%
	\begin{algorithmic}[1]
		\STATE \textbf{Input:} number of particles $N$, unnormalized annealed targets $\{\unnormDensity_k\}_{k=0}^K$ such that $\unnormDensity_0=\targetDist_0$ and $\unnormDensity_K=\unnormDensity$, resampling threshold $A\in\left[1/N,1\right)$.
		\STATE Output: Approximations $\targetDist^N_K$ and $\normConst_K^N$ of $\targetDist$ and $\normConst$.%
		\STATE Sample $X^i_0 \sim \targetDist_0$ and set $W_0^i = \frac{1}{N}$ and  $\normConst_0^N=1$.
		\FOR{$k=1,\dots, K$}
			\STATE Solve
				$\baseFlow_k  \leftarrow \text{argmin}_{T \in \mathcal{T}}~\mathcal{L}^N_k(T)$ using e.g. SGD.
						\STATE $\left(\targetDist^\numParticles_\timeIndex,\normConst_\timeIndex^\numParticles \right) {\leftarrow} \verb+ SMC-NF-step+\left(\targetDist^\numParticles_{\timeIndex-1},\normConst_{\timeIndex-1}^\numParticles, \baseFlow_\timeIndex\right)$
		\ENDFOR
	\end{algorithmic}
\end{algorithm}

\begin{algorithm}
\caption{CRAFT to simple AFT line difference: additions for CRAFT shown in \textcolor{green}{green} and removals from simple AFT shown in \textcolor{red}{\sout{red}}}\label{algo:CRAFTdiff}
	\begin{algorithmic}[1]
		\STATE \textbf{Input:} number of particles $N$, unnormalized annealed targets $\{\unnormDensity_k\}_{k=0}^K$ such that $\unnormDensity_0=\targetDist_0$ and $\unnormDensity_K=\unnormDensity$, resampling threshold $A\in\left[1/N,1\right)$.
		\STATE Output: \textcolor{green}{ Length $J$ sequence of} approximations $\targetDist^N_K$ and $\normConst_K^N$ of $\targetDist$ and $\normConst$.%
		\FOR{\textcolor{green}{$j=1,\dots, J$}}
		\STATE Sample $X^i_0 \sim \targetDist_0$ and set $W_0^i = \frac{1}{N}$ and  $\normConst_0^N=1$.
		\FOR{$k=1,\dots, K$}
			\STATE \textcolor{red}{\sout{Solve
				$\baseFlow_k  \leftarrow \text{argmin}_{T \in \mathcal{T}}~\mathcal{L}^N_k(T)$ using e.g. SGD.}}
            \STATE $\textcolor{green}{\hat{h} \leftarrow} {\color{green}\verb+flow-grad+\left(\baseFlow_\timeIndex, \targetDist_{\timeIndex-1}^\numParticles \right)}$ \textcolor{green}{using eqn (8).}
			\STATE $\left(\targetDist^\numParticles_\timeIndex,\normConst_\timeIndex^\numParticles \right) {\leftarrow} \verb+ SMC-NF-step+\left(\targetDist^\numParticles_{\timeIndex-1},\normConst_{\timeIndex-1}^\numParticles, \baseFlow_\timeIndex\right)$
			\STATE \textcolor{green}{Apply gradient update $\hat{h}$ to flow $\baseFlow_k$ }
			
		\ENDFOR
		\STATE \textcolor{green}{Yield 	 $(\targetDist^\numParticles_{\numTransitions},\normConst_{\numTransitions}^\numParticles)$ and continue for loop.}
	\ENDFOR
	\end{algorithmic}
\end{algorithm}

\begin{algorithm}
\caption{Practical AFT: from Arbel, Matthews and Doucet 2021. This corresponds to Algorithm 2 of the prior paper.}\label{algo:AFTpractical} 	%
	\begin{algorithmic}[1]
		\STATE \textbf{Input:} number of training, test and validation particles $N_{\textup{train}}$, $N_{\textup{test}}$, $N_{\textup{val}}$, unnormalized annealed targets $\{\gamma_k\}_{k=0}^K$ such that $\gamma_0=\pi_0$ and $\gamma_K=\gamma$, resampling thresholds $A_a\in\left[1/{N_a},1\right)$ for $a \in \{\textup{train},\textup{test},\textup{val}\}$, number of training iterations $J$.
		\STATE \textbf{Output:} Approximations $\pi^{N_{\textup{test}}}_{K, \textup{test}}$ and $Z_{K, \textup{test}}^{N_{\textup{test}}}$ of $\pi$ and $Z$.
		\FOR{$a \in \{\textup{train},\textup{test},\textup{val}\}$}
			\STATE Sample  $X^{i,a}_0 \sim \pi_0$ and set $W_0^{i,a} \leftarrow \frac{1}{N_a}$ and  $Z_0^{N,a}\leftarrow 1$.
		\ENDFOR
		\FOR{$k=1,\dots, K$} 
			\STATE Learn the flow
				$T_k  \leftarrow \verb+EarlyStopLearnFlow+\parens{J,X^{N_{\textup{train}}}_{k-1, \textup{train}} ,W^{N_{\textup{train}}}_{k-1, \textup{train}},  X^{N_\textup{val}}_{k-1, \textup{val}}, W^{N_\textup{val}}_{k-1, \textup{val}}}$
			\FOR{$a \in \{\textup{train},\textup{test},\textup{val}\}$}
						\STATE $\left(\targetDist^{\numParticles_{a}}_{\timeIndex,a},\normConst_{\timeIndex, a}^{\numParticles_a} \right) {\leftarrow} \verb+ SMC-NF-step+\left(\targetDist^{\numParticles_{a}}_{\timeIndex-1, a},\normConst_{\timeIndex-1, a}^{\numParticles_{a}}, \baseFlow_\timeIndex\right)$			
			\ENDFOR
		\ENDFOR
	\end{algorithmic}
\end{algorithm}

\begin{algorithm}
\caption{Subroutine \emph{EarlyStopLearnFlow} for practical AFT.}	%
	\begin{algorithmic}[1]
		\STATE \textbf{Input:} Number of training iterations $J$, training and validation particles and weights $\braces{X^{i,\textup{train}}_{k-1} , W^{i, \textup{train}}_{k-1} }_{i=1}^{N_{\textup{train}}}$ and  $\braces{X^{i,\textup{val}}_{k-1} , W^{i, \textup{val}}_{k-1} }_{i=1}^{N_{\textup{val}}}$.
		\STATE \textbf{Ouput:} Estimated flow $T_k$
		\STATE Initialize flow to identity $T_k= I_D$.
		\STATE Initialize list of flows $\mathcal{T}_{opt} \leftarrow \{T_k\}$.
		\STATE Initialize list of validation losses\\
		    $\mathcal{E}\leftarrow  \braces{\sum_{i=1}^{N_{\textup{val}}} W_{k-1}^{i,\textup{val}} D_k\parens{X_{k-1}^{i,\textup{val}}}}$
		\FOR{$j=1,...,J$}
		    \STATE Compute training loss using \eqref{eq:lHatAlgo}  \\
		        $ \mathcal{L}_k^{N_{\textup{train}}}(T_k) \leftarrow \sum_{i=1}^{N_{\textup{train}}} W_{k-1}^{i,\textup{train}} D_{k}\parens{X_{k-1}^{i,\textup{train}}}$.
		    \STATE Update $T_k$ using SGD step on $\mathcal{L}_k^{N_{\textup{train}}}(T_k)$ with approx. gradient \eqref{eq:gradientApproximationAlgo}.
		    \STATE Update list of flows $\mathcal{T}_{opt} \leftarrow \mathcal{T}_{opt}\cup \{T_k\}$
		    \STATE Update list of validation losses $\mathcal{E}$\\
		    $\mathcal{E}\leftarrow  \mathcal{E}\cup \braces{\sum_{i=1}^{N_{\textup{val}}} W_{k-1}^{i,\textup{val}} D_{k}\parens{X_{k-1}^{i,\textup{val}}}}$
		\ENDFOR
	    \STATE Return flow with smallest validation error from the list of flows $\mathcal{T}_{opt}$.
	\end{algorithmic}
\end{algorithm}

\end{document}